%% file: arxiv_version.tex
\documentclass[a4paper]{article}
\usepackage{jmlr2e} 
\usepackage{amsmath,graphicx}

\usepackage{graphics, theorem, times, amsfonts, amsmath, amssymb, cite}
\usepackage{tikz}
\usetikzlibrary{shapes,arrows}
\usepackage{pgfplots}
\usepackage{color}
\input{my_sections}

\usepackage{hyperref}
\usepackage{multirow}
\usepackage{rotating}
\usepackage{comment}
\usepackage{flushend}
\usepackage{caption}
\usepackage{subcaption}
\usepackage[affil-it]{authblk}
\usepackage{setspace}

\usepackage{endnotes}

\usepackage{titlesec}

\usepackage{har2nat}
\spacing{1.5}

\input{mysymbol.sty}

\titleformat{\section}    
       {\normalfont\fontfamily{phv}\fontsize{12}{17}\bfseries}{\thesection}{1em}{}

\titleformat{\subsection}    
       {\normalfont\fontfamily{phv}\fontsize{12}{17}\itshape}{\thesubsection}{1em}{}

\usepackage{morefloats}

\begin{document}

\title{Stylometric Analysis of Early Modern Period English Plays}
%
\author[1]{Mark Eisen}
\author[2]{Santiago Segarra}
\author[3]{Gabriel Egan}
\author[1]{Alejandro Ribeiro}
\affil[1]{Dept. of Electrical and Systems Engineering\\ University of Pennsylvania, Philadelphia, USA}
\affil[2]{Inst. for Data, Systems, and Society, Massachusetts Institute of Technology, Cambridge, USA}
\affil[3]{School of Humanities\\ De Montfort University, Leicester, UK}
\editor{}

\maketitle

\begin{abstract}

Function word adjacency networks (WANs) are used to study the authorship of plays from the Early Modern English period. In these networks, nodes are function words and directed edges between two nodes represent the relative frequency of directed co-appearance of the two words. For every analyzed play, a WAN is constructed and these are aggregated to generate author profile networks. We first study the similarity of writing styles between Early English playwrights by comparing the profile WANs. The accuracy of using WANs for authorship attribution is then demonstrated by attributing known plays among six popular playwrights. Moreover, the WAN method is shown to outperform other frequency-based methods on attributing Early English plays. In addition, WANs are shown to be reliable classifiers even when attributing collaborative plays. For several plays of disputed co-authorship, a deeper analysis is performed by attributing every act and scene separately, in which we both corroborate existing breakdowns and provide evidence of new assignments. 

\end{abstract}
%


\section{Introduction}\label{sec_introduction}

Stylometry involves the quantitative analysis of a text's linguistic features in order to gain further insight into its underlying elements, such as authorship or genre. Along with common uses in digital forensics \citep{de2001mining, stamatatos2009survey} and plagiarism detection \citep{meuschke2013state}, stylometry has also become the primary method for evaluating authorship disputes in historical texts, such as the Federalist papers  (Mosteller and Wallace, 1964; Holmes and Forsyth, 1995) 
and the Mormon scripture \citep{holmes1992stylometric}, in a field called authorship attribution. Such disputes exist regarding the collection of dramatic works produced in England during the Early Modern era, covering the 16th through mid-17th century. Due to factors such as inaccurate publication information on title pages and undocumented collaborations, the precise authorship of many of these plays--including works by William Shakespeare and John Fletcher--remains highly contested. 

Stylometric analysis of the work from this time period dates as far back as the nineteenth century in F. G. Fleay's analysis of verse features in Shakespeare's plays \citep{fleay1878shakespeare}. Similar analyses based on the manual counting of linguistic features continued throughout the early to late twentieth century (Timberlake, 1931; Oras, 1960; Tarlinskaja et al., 1987). 
Computer-based techniques for counting the frequency of various stylistic features, such as rare words or phrases, have become very common over the past few decades. The most recent work done in evaluating authorship in Early Modern era drama includes that by MacDonald P. Jackson \citep{jackson2003defining, jackson2006shakespeare}, Brian Vickers \citep{vickers2002shakespeare}, and Hugh Craig and Arthur Kinney \citep{craig2009shakespeare}, each of whom studied the works of Shakespeare and his contemporaries extensively using computational stylometry techniques. 

The techniques used in modern authorship attribution began almost a century ago by examining sentence lengths in texts to determine authorship \citep{yule1939sentence}. \citet{mosteller1964inference} were the first to consider function words as important stylistic markers in stylometric analysis, producing unprecedented results. As such, function words have continued to be common in analysis techniques \citep{argamon2005measuring, juola2006authorship} due to their context independence and ubiquity at high rates of occurence in English language texts. These methods rely mainly on the frequency of usage of function words. Numerous other stylistic features have since been used in authorship attribution studies, including  vocabulary richness \citep{holmes1991vocabulary, hoover2003another} and parts-of-speech \citep{cutting1992practical}.

Our method for attributing texts, developed in \citep{DBLP:journals/corr/SegarraER14}, also measures function word usage to distinguish author styles. Rather than only considering word frequencies, however, we consider a more complex relational structure in an author's usage of function words. We construct word adjacency networks (WANs) with function words as nodes, and edges containing information regarding the use of two function words within a certain distance (measured in intervening words) from one another. We interpret each WAN as a Markov chain that assigns transition probabilities to the appearance of two function words in succession, derived from their actual occurrences in succession at varying distances within the securely attributed texts. Thus, these probabilities stand for the author's expressed preference for following one particular word with another. We can then quantify similarity between WANs by using a measure of relative entropy. Markov chains have previously been used in \citep{khmelev2001using} and \citep{sanderson2006short} for the purposes of authorship attribution, though neither consider the use of function words. Results in \citep{DBLP:journals/corr/SegarraER14} show an increase in attribution accuracy compared to frequency-based methods for general texts of English literature. In this work we perform further validation of the method's performance specifically on plays from the Early Modern period and compare this performance to that of word frequency-based methods previously used in Shakespeare attributional studies. We then employ this new technique to comment on authorship disputes concerning Early Modern English dramatic works.

We first present an overview of the construction and comparison of WANs in Section \ref{sec_word_adjacency_network}. We discuss in Section \ref{sec_author_profiles} the main playwrights used in our analysis as well as the construction of their profile networks, and in Section~\ref{sec_profile_similarity} we present a measure of similarity between profiles. As a validation of the method, in Section \ref{sec_attribution_of_plays} we perform a stylometric analysis of the complete undisputed works of our six primary playwrights, followed by a comparison with existing methods in Section \ref{sec_summary}. We are able to demonstrate high attribution accuracy in discriminating between six candidate authors. We then examine the use of WANs in determining authorship of plays known to be written by multiple authors in collaboration. This is first done by analyzing entire plays in Section \ref{sec_collaborations} and then through extensive interplay analysis of a set of particularly controversial plays in Section \ref{sec_collab_intraplay}. Our results largely corroborate existing theories regarding these plays and, in some cases, propose new divisions of labor.

\section{Word Adjacency Networks}\label{sec_word_adjacency_network}

When doing authorship attribution, we are given a set of candidate authors $A = \{a_1, a_2, \ldots, a_n\}$ and a set of known texts written by each of these authors, and the objective is to correctly attribute a collection of texts of unknown authorship among the authors. 
In \citep{DBLP:journals/corr/SegarraER14, segarra2013authorship}, we propose an authorship attribution method based on function word adjacency networks. For each text, we can construct a word adjacency network (WAN) of function words. These include prepositions, conjunctions, pronouns, auxiliary verbs, and articles that convey only grammatical relationships between the so-called lexical words that carry meaning. Formally, from a given text $t$ we construct the network $W_t = (F,Q_{t})$ where $F = \left\{f_{1},f_{2},...,f_{f}\right\}$ is the set of nodes composed by a collection of
function words and $Q_{t}$ is a similarity measure between ordered pairs of function words. 

The similarity function $Q_t$ measures the directed co-appearance of two function words. Once we encounter a particular function word, $Q_t$ indicates the likelihood of encountering another one in the few words following the first one.  
More precisely, to compute $Q_{t}$ we first divide the text $t$ into units of consecutive words (e.g. sentences, speeches) $s^{h}_{t}$ where $h$ ranges from 1 to the total number of units. We denote by $\ s^{h}_{t}(e)$ the word in the$\ e$-th position within unit $h$ of text $t$. Moreover, we consider that two words in the same unit are related if they are at most $D \in \naturals$ positions apart and the relation between words decays with their position difference according to a discount factor $\alpha \in (0, 1)$. In this way, with $\mathbb{I}\{ \cdot \}$ denoting the indicator function, we define 
\begin{equation} \label{eqn_definition_similarity_function_text}
Q_{t}(f_{i},f_{j}) \! = \! \sum_{h,e} \mathbb{I}\left\{ s^{h}_{t}(e) = f_{i}\right\} \sum_{d=1}^{D} \alpha^{d-1} \,\, \mathbb{I}\left\{ s^{h}_{t}(e + d) = f_{j}\right\}\!,
\end{equation}
for all$\ f_{i}, f_{j} \in F$. The selection of the decay parameter $\alpha$, the window size $D$, and the delimiting units $s^h_t$, in general, may vary based on the texts and authors being considered. In this work, we select $\alpha=0.75$ and $D=10$, determined in \citep{DBLP:journals/corr/SegarraER14} to be generally optimal and robust parameter choices. However, because punctuation marks were often added by publishers rather than the authors themselves \citep{howard1930printer}, and because dramatic characters do not necessarily speak in sentences, when applying our method to Early Modern plays (rather than novels) we use individual speeches (rather than clauses or sentences) as the units into which we break our texts.

We then generate a profile network $W_{c}= (F, Q_{c})$ for every author $a_c$ using the WANs from those texts known to have been written by the corresponding author $a_c$. Formally, if we denote by $T^{(c)}$ the set of texts written by author $a_c$, then the similarity function $Q_c$ of the profile is computed as 
\begin{equation} \label{eqn_definition_similarity_function_author}
Q_{c} = \sum_{t \in T^{(c)}} Q_{t}.
\end{equation}

The similarity function $Q_{c}$ depends on the number and length of the texts written
by author $a_{c}$. This is a problem since we aim to compare profiles of different authors whose canons will be of differing sizes. Thus, we apply the following normalization to the similarity measures
\begin{equation} \label{eqn_definition_normalization_similarity_function}
\hat{Q}_c(f_i, f_j) = \frac{Q_c(f_i,f_j)}{\sum_{j} Q_c(f_i,f_j)},
\end{equation}
for all $f_i, f_j \in F$. In \eqref{eqn_definition_normalization_similarity_function} we assume that the combined length of the texts written by author $a_c$ is long enough to guarantee a non-zero denominator for a given number of function words $|F|$. If this is not the case for some function word $f_i$, we fix $\hat{Q}_c(f_i, f_j) = 1/|F|$ for all $f_j$. In this way, we achieve normalized networks $P_c=(F, \hat{Q}_c)$ for each author $a_c$. The network $P_c$ provides an estimate of the potentially discriminative word selection preferences  of author $a_c$.
Since the similarities out of every node sum up to 1 in the network $P_c$, it can be interpreted as a discrete time Markov chain (MC). Thus, the normalized similarity $\hat{Q}_c(f_i, f_j)$ between words $f_i$ and $f_j$ is a measure of the probability of finding $f_j$ in the words following an encounter of $f_i$ for texts written by author $a_c$.
Similarly, we can use normalization \eqref{eqn_definition_normalization_similarity_function} to build a MC $P_{u}$ for each unknown text.

In order to perform the attribution, we need a way of comparing the generated MCs. By construction, every MC has the same state space $F$, facilitating the comparison. Indeed, we use the relative entropy $H(P_1, P_2)$ as a dissimilarity measure between any two chains $P_1$ and $P_2$. The relative entropy is given by
\begin{equation}\label{eqn_definition_relative_entropy}
H(P_1, P_2)= \sum_{i,j} \pi(f_i) P_1(f_i, f_j) \log \frac{P_1(f_i, f_j)}{P_2(f_i, f_j)},
\end{equation}
where $\pi$ is the limiting distribution of $P_1$ and we consider $0 \, \log 0$ to be equal to $0$. From \citep{kesidis1993relative}, if we denote as $w_1$ a realization of the MC $P_1$ then $H(P_1, P_2)$ is proportional to the logarithm of the ratio between the probability that $w_1$ is a realization of $P_1$ and the probability that $w_1$ is a realization of $P_2$. In particular, when $H(P_1, P_2)$ is null, the ratio of probabilities is 1, meaning that a given realization of $P_1$ has the same probability of being observed in both MCs. Thus, $H$ is a reasonable dissimilarity measure between MCs. Utilizing \eqref{eqn_definition_relative_entropy} we construct the attribution function $\hat{r}_{U}$ by assigning the text $u$ to the author with the MC most similar to $P_u$, i.e. 
\begin{equation}\label{eqn_def_r_hat_T_0}
\hat{r}_{U}(u)= a_{c^*}, \,\,\, \text{where} \,\,\, c^*=\argmin_c H(P_{u}, P_c).
\end{equation}

Notice that the relative entropy in \eqref{eqn_def_r_hat_T_0} takes an infinite value when any word-pair collocation that appears in the unknown text does not appear in the profile. In practice we compute the relative entropy in \eqref{eqn_definition_relative_entropy} by summing only over the non-zero transitions in the profiles, 
\begin{equation}\label{eqn_def_H_in_practice}
H(P_1, P_2)= \sum_{i,j | P_2(f_i, f_j) \neq 0 } \pi(f_i) P_1(f_i, f_j) \log \frac{P_1(f_i, f_j)}{P_2(f_i, f_j)}.
\end{equation}
Because the calculation of relative entropy in \eqref{eqn_def_H_in_practice} only adds relative entropy for nonzero transitions, profiles built from fewer total words will on average contain fewer non-zero transitions and will thus sum together fewer terms than larger profiles. When attributing an unknown text among profiles of differing sizes, we avoid this potential biasing for smaller profiles by  summing only over transitions that are non-zero in every profile being considered,
\begin{equation}\label{eqn_def_H_in_practice_global}
H(P_1, P_2)= \sum_{\substack{i,j |P_c(f_i, f_j) \neq 0 \\ \forall \,\, a_{c} \in A}} \pi(f_i) P_1(f_i, f_j) \log \frac{P_1(f_i, f_j)}{P_2(f_i, f_j)}.
\end{equation}
In the following sections, expression \eqref{eqn_def_H_in_practice_global} is used to compare the Markov chain representations of WANs when performing attributions following rule \eqref{eqn_def_r_hat_T_0}.

\section{Author Profiles} \label{sec_author_profiles}

The stylometric analysis in this paper focuses on the attribution of plays written during the English Early Modern period stretching from the late 16th century to the early 17th century. William Shakespeare is the most prominent playwright active in this period but there are several other authors that were also active during this time. For most of the paper, we focus on just six playwrights, whose life-spans (in brackets) and assumed career-spans are:\footnote{Information compiled from the Database of Early English Playbooks (DEEP) \citep{DEEP} and the database of catalogued plays in Literature Online (LION) \citep{LION}. Whenever inconsistencies in authorship information arise, we accept the verdicts in \citep{DEEP} unless compelling new research contradicts them.}

\begin{enumerate}
\item[(1)] George Chapman (1559-1634), active circa 1596-1620. 
\item[(2)] Christopher Marlowe (1564-1593), active circa 1586-1593. 
\item[(3)] William Shakespeare (1564-1616), active circa 1589-1614.
\item[(4)] Ben Jonson (1572-1637), active circa 1596-1637.
\item[(5)] John Fletcher (1579-1625), active circa 1605-1625.  
\item[(6)] Thomas Middleton (1580-1627), active circa 1603-1625. 
\end{enumerate}

\noindent We focus on plays for the professional public theatre and disregard works commissioned for one-off and/or private performance such as masques, entertainments and pageants. Chapman, Marlowe, Shakespeare, Jonson, Fletcher, and Middleton are central to our analysis since they created large and well-studied canons compared to their contemporaries.

The WAN attribution algorithm developed in \citep{DBLP:journals/corr/SegarraER14} and briefly reviewed in Section \ref{sec_word_adjacency_network} uses known texts of a given author to construct a profile against which unknown texts are compared. Since profiling accuracy increases with the length of the texts used in building the profile, we use all  texts of sole authorship that have little or no history of authorship dispute. The full list of plays used to build the six profiles is reported in Table \ref{table_profile_texts} in the Appendix. When building profiles for a given author, we generally use the DEEP database \citep{DEEP} to determine texts of sole authorship. 
An exception to this is Middleton, for whom the 2007 Oxford Collected Works of Middleton \citep{taylor2007thomas} is taken as the most reliable source.

Notice that each profile is built from a different number of texts. Marlowe, the least prolific writer of the ones here considered, is accepted as the sole author of 6 plays containing a total of 103,160 words. Shakespeare, the most prolific writer, is the undisputed sole author of 28 plays, containing 679,256 words. Due to this difference, we compute the relative entropy between the WAN of an unknown text and each profile using \eqref{eqn_def_H_in_practice_global} rather than \eqref{eqn_def_H_in_practice}.

In order to prevent distortions introduced by different editions handling modernization differently--Shakespeare typically being more heavily modernized than other writers--we rely on  the earliest editions available of each text in the LION database \citep{LION}, with the exception of Shakespeare plays for which multiple early editions exist. About half of Shakespeare's plays were first published during his lifetime in single-play editions known (from their paper format) as quartos, and some of these plays went through multiple quarto editions. Seven years after his death, a collection of 36 of his plays was published in a book now known as the First Folio (1623), forming the foundation of his canon. Thus for many plays we have multiple quarto editions and the Folio edition to choose from, and in most cases scholars have reached no general consensus about which of these editions best reflect Shakespeare's own intentions for his works. Because the First Folio edition was manufactured by one team of workmen in one printshop over a relatively short period of time (1622-23), we choose, in those cases where there is a choice of editions to be made, the Folio text over any preceding quarto. When using original transcriptions we have to account for the fact that many words had multiple accepted spellings during the Early Modern era. In general, spelling preferences in printed editions are a poor guide to authorship because printers were free to alter spellings whenever doing so assisted in producing fully justified lines of type \citep{gaskell1972new}. However, the alternative spellings occur infrequently (relative to the high frequency of function words, in general) and do not affect the conclusions made by our method, and are therefore ignored. In addition, we remove speech prefixes, meaning the character name preceding each speech, to avoid cases in which character names are abbreviated to function words (such as Anne abbreviated to `An').

The WANs for each play and author profile are built using up to 100 of the most common function words from the Early Modern period, listed in Table \ref{table_function_words_all} in the Appendix. The number of function words chosen from the full set of 100 varies for each experiment and is determined by a training process in which we measure the power of each word in helping to discriminate between the particular authors under consideration.
%


\section{Similarity of Profiles}\label{sec_profile_similarity}

\begin{table}
\caption{Relative entropy between profiles.} \label{table_profile_comparison}
\renewcommand{\arraystretch}{2.5}
\centering
\resizebox{0.5\textwidth}{!}{\input{table_profile_distances.tex}}
\end{table}

\begin{figure}
\centering
\includegraphics[height=.23\textheight]{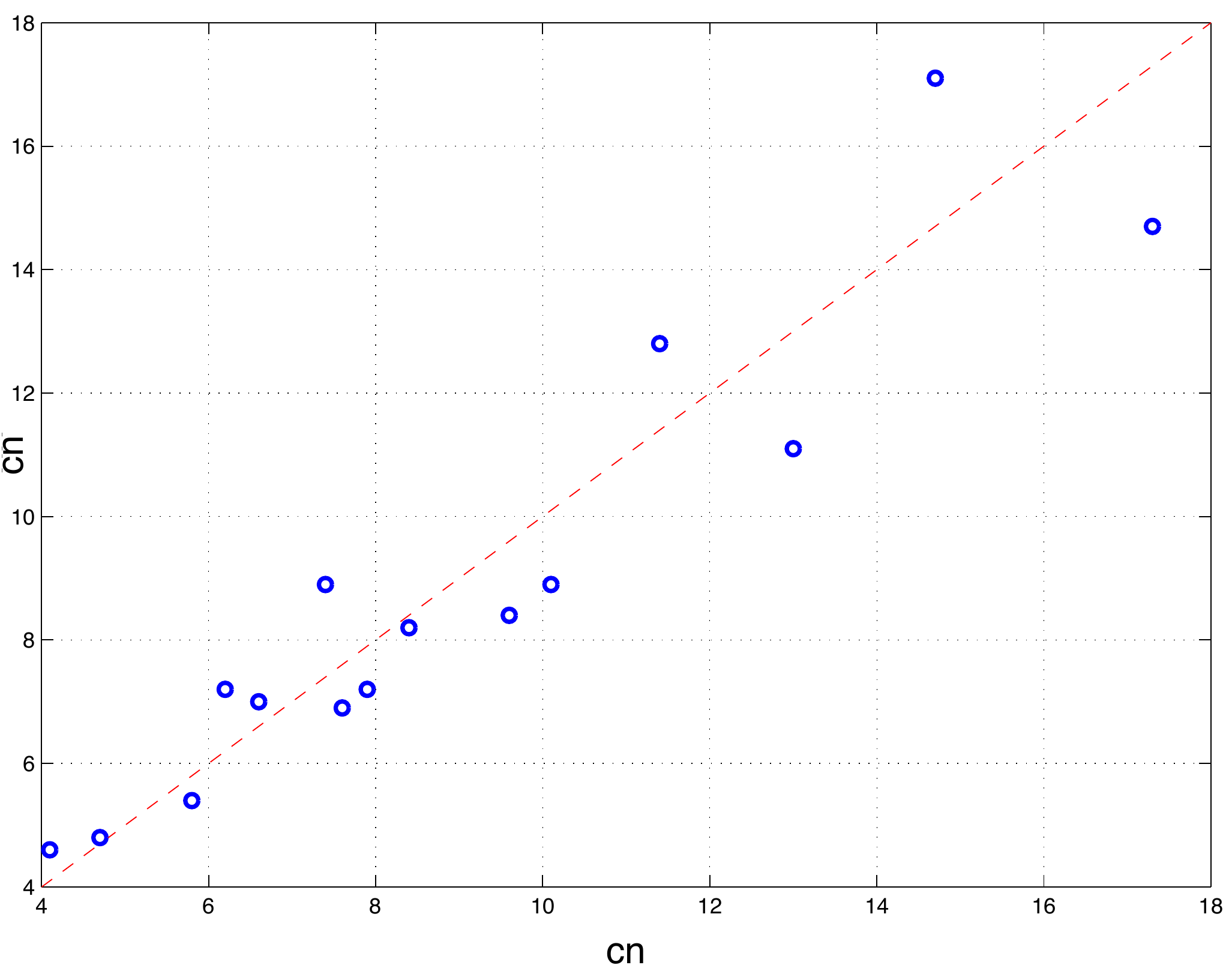}
\caption{Asymmetry of dissimilarities in Table \ref{table_profile_comparison}.}
\label{fig_symmetry_profiles}
\end{figure}

We compute the relative entropy between every pair of author profiles for the six authors introduced in Section \ref{sec_author_profiles} using expression \eqref{eqn_def_H_in_practice_global}; see Table \ref{table_profile_comparison}. Every entry in the table represents the relative entropy between the corresponding authors in the rows and columns. In this table, as well as in the remaining of the paper, relative entropies are multiplied by 100 to scale the figures up and thereby give results that are more easily compared by eye. We use the term centinats, or $cn$ for short, to denote the resultant unit of measure of relative entropy. The 4.7 in the Chapman row entry and Shakespeare column entry indicates a relative entropy of $4.7cn$ between Chapman's and Shakespeare's profiles. Note that, as expression \eqref{eqn_def_H_in_practice_global} is not symmetric, the values in the table are asymmetric, although they are similar in most cases. Thus the relative entropy between Shakespeare's and Chapman's profiles is $4.8cn$ when Shakespeare's profile is taken first and Chapman's second, but only $4.7cn$ when Chapman's profile is taken first and Shakespeare's second. This is an inevitable consequence of the comparison method's non-commutativity.
In general, dissimilarities between profiles in both directions are highly correlated as can be observed in Fig. \ref{fig_symmetry_profiles}. In this figure, the coordinates of every point correspond to the dissimilarities in both directions for every pair of profiles. The arrangement of the points along the diagonal implies that a high dissimilarity in one direction is associated with a high dissimilarity in the opposite one. Hence, this correlation allows us to speak about the similarity between two authors without specifying a direction.

The entropy-based dissimilarities in Table \ref{table_profile_comparison} dispel the Marlovian theory of Shakespeare authorship \citep{webster1923marlowe}. If Marlowe wrote the works traditionally attributed to Shakespeare, we should observe the dissimilarities between Marlowe's and Shakespeare's profile to be smaller than the distances between each of the other profiles. However, the relative entropies between Marlowe's and Shakespeare's profiles average $9.5cn$ in both directions which is larger than the dissimilarity between Shakespeare and all of the other authors. Shakespeare's profile is, on average, closest to Jonson profile -- average relative entropy of $4.4cn$ -- although still sufficiently different that we can be sure that these are not two names for the same man, as verified by the attribution of plays in Section \ref{sec_attribution_of_plays}. The highest dissimilarity among any pair of profiles occurs between Marlowe and Fletcher with a mean of $16.1cn$. As will be seen in Section \ref{sec_attribution_of_plays}, the relative similarity between two profiles affects our ability to distinguish between them when attributing a text.

\section{Attribution of Plays}\label{sec_attribution_of_plays}

As a means of validating the accuracy of the WAN method on Early Modern English dramas, we first use it to attribute the undisputed works of Jonson, Middleton, Chapman, Marlowe, Shakespeare, and Fletcher among these six authors.. When attributing any given play, profiles are built using the plays listed in Table \ref{table_profile_texts} excluding the one being attributed. We do not report raw relative entropy values between the play being attributed and the author profiles, but instead subtract from these values the relative entropy between the play and a profile containing all available texts. Intuitively, the profile containing all of the texts represents the writing style of a hypothetical average playwright from this period. This is done to make the figures easier to view but does not change the results in any way. Each raw relative entropy value is discounted by the same constant value, thus preserving relative dissimilarities. As a result, both negative and positive relative entropy values are possible. A negative relative entropy value indicates that the play's WAN is more similar to the author profile than to the profile of the average playwright while a positive relative entropy indicates the opposite.


\begin{figure*}
\centering
\includegraphics[height=.3\textheight, width = 0.82\textwidth]{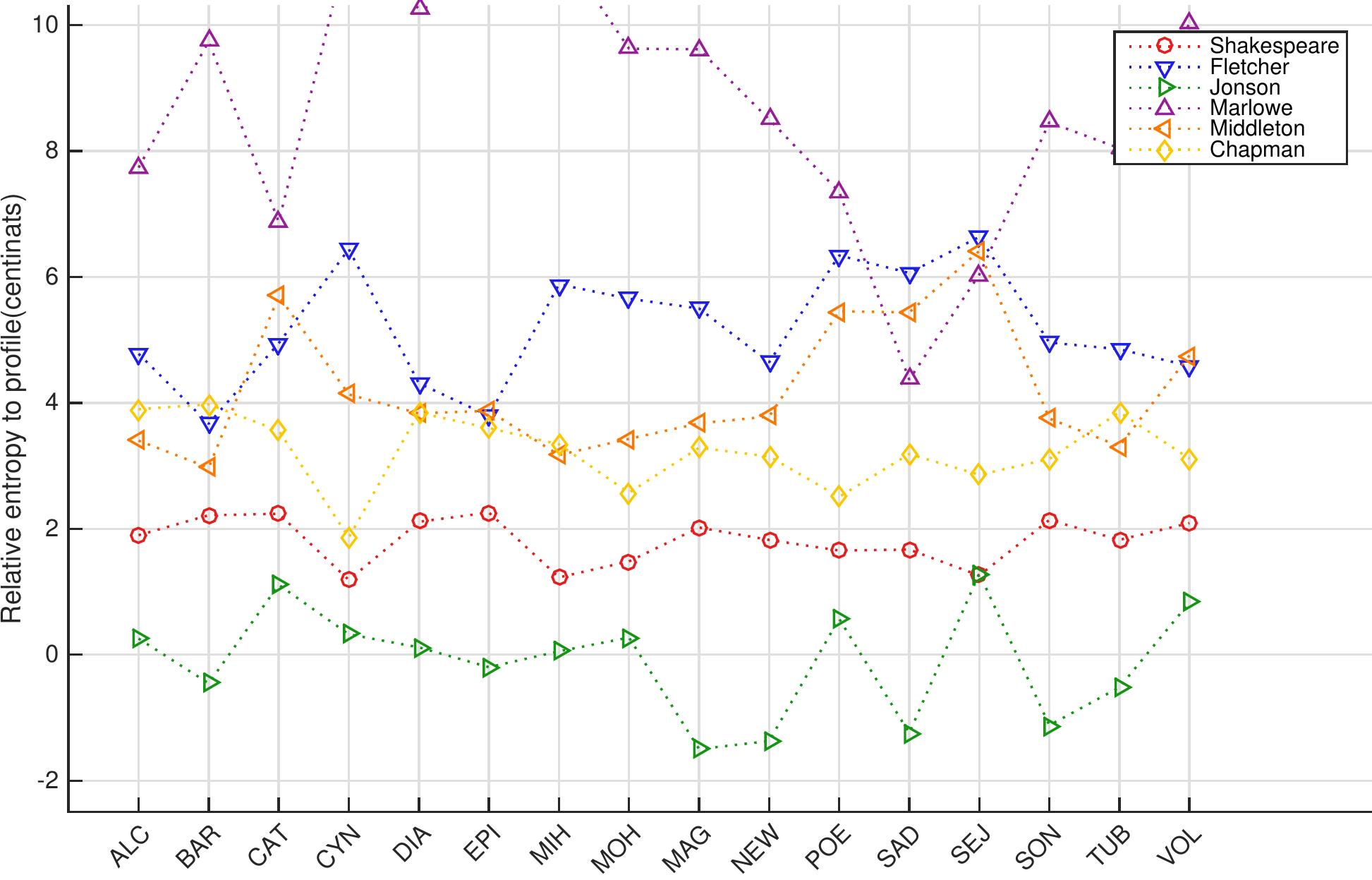}
\caption{Attribution of Jonson plays. The 16 known-to-be Jonson plays in Table \ref{table_profile_texts} are all correctly attributed to Jonson by our method.}
\label{fig_jonson}
\end{figure*}

In Fig. \ref{fig_jonson} we present our method's attribution of the 16 plays known to have been written by Ben Jonson in Table \ref{table_profile_texts}. In the horizontal axis we present the plays to attribute and the vertical axis represents the relative entropy \eqref{eqn_def_H_in_practice_global} in $cn$ from these plays to the different profiles identified with distinct markers and discounted by the distance to the average playwright. Observe that we achieve 100\% accuracy in attributing the works of Ben Jonson. Note that the play \emph{Sejanus His Fall} is virtually a tie between Jonson and Shakespeare.

\begin{figure*}
\centering
\includegraphics[height=.3\textheight, width = 0.82\textwidth]{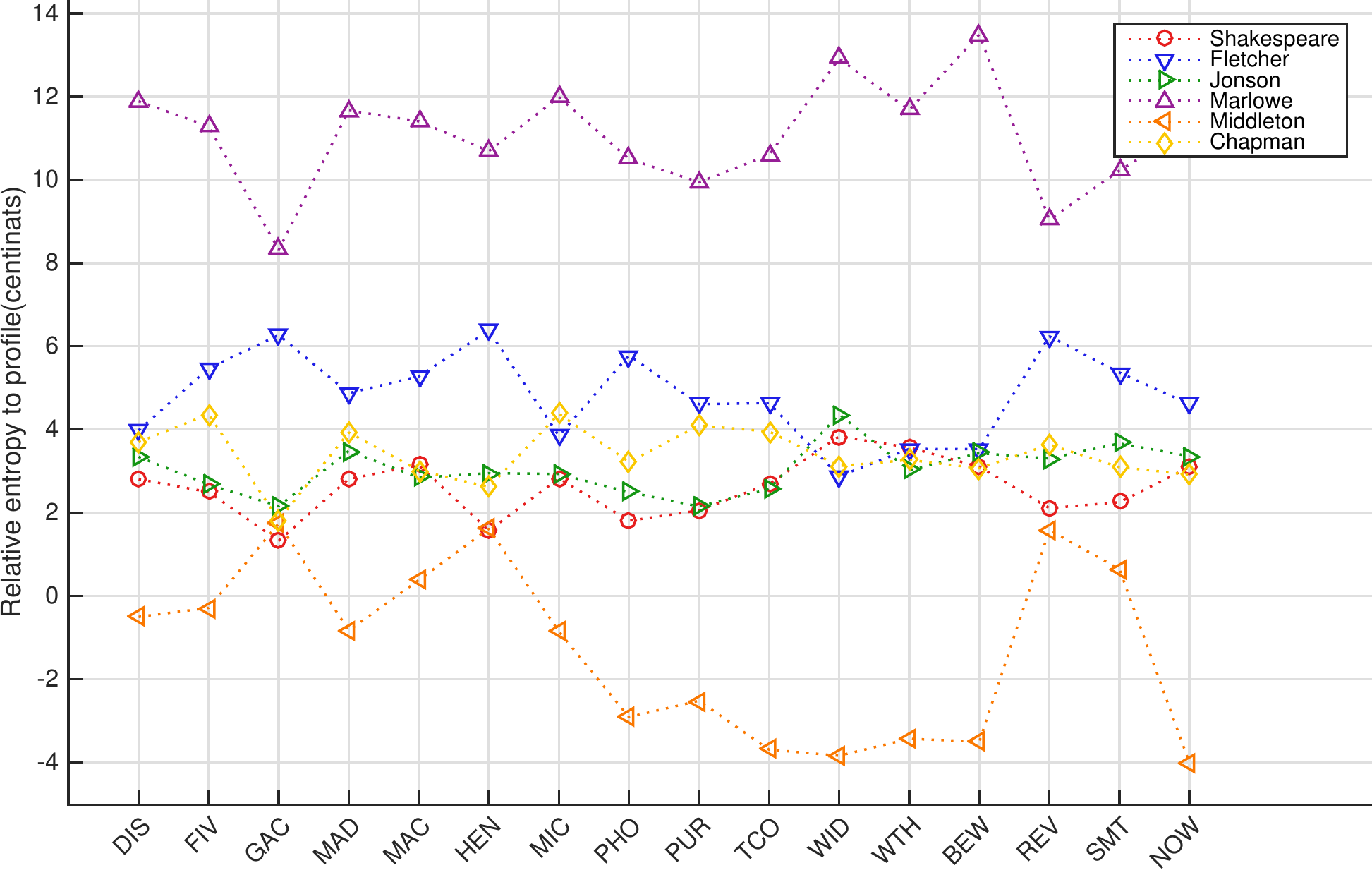}
\caption{Attribution of Middleton plays. Of the 16 known-to-be Middleton plays in Table \ref{table_profile_texts}, two sole-authored plays are misattributed by our method.}
\label{fig_middleton}
\end{figure*}

In Fig. \ref{fig_middleton} we present the attribution of 16 plays written by Thomas Middleton, with 14 correctly assigned to Middleton by our method. The first misattributed play, \emph{A Game at Chess}, is attributed to Shakespeare by a very small margin, likely due to random error. This is also true in the case of \emph{Hengist King of Kent}, noted for being the only history play Middleton wrote.

\begin{figure}
\centering
\includegraphics[height=.3\textheight, width=0.8\textwidth]{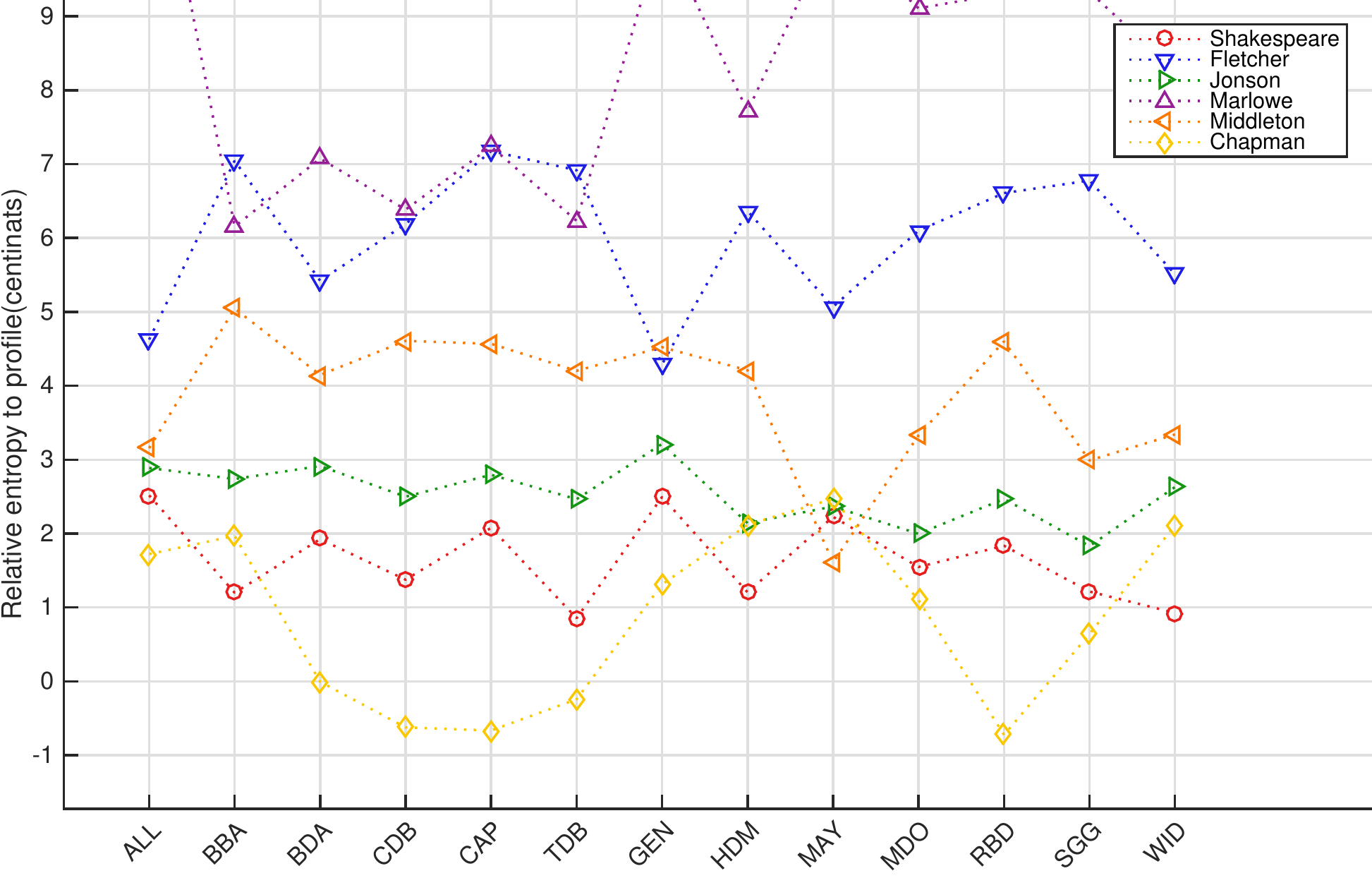}
\caption{Attribution of Chapman plays. Of the 13 plays known-to-be Chapman plays in Table \ref{table_profile_texts}, four sole-authored plays are misattributed by our method.}
\label{fig_chapman}
\end{figure}

George Chapman is widely accepted as the author of 13 plays listed in Table \ref{table_profile_texts} and attributed by our method in Fig. \ref{fig_chapman}. In total, 9 of the 13 plays are correctly attributed to Chapman by our method. Of the misattributions, three are assigned to Shakespeare with Chapman as the second nearest candidate. This is consistent with the fact that in Table \ref{table_profile_comparison}, Chapman's profile is similar to Shakespeare and hence they are difficult to distinguish with our method. Thus, cases of random error will most likely make our method attribute to Shakespeare plays that were in fact written by Chapman. We are fortunate, however, that theatre history gives us no reason to suppose that they ever collaborated, and in practice there are no significant authorship disputes that involve both of them.


\begin{figure}
\centering
\includegraphics[height=.3\textheight, width=0.8\textwidth]{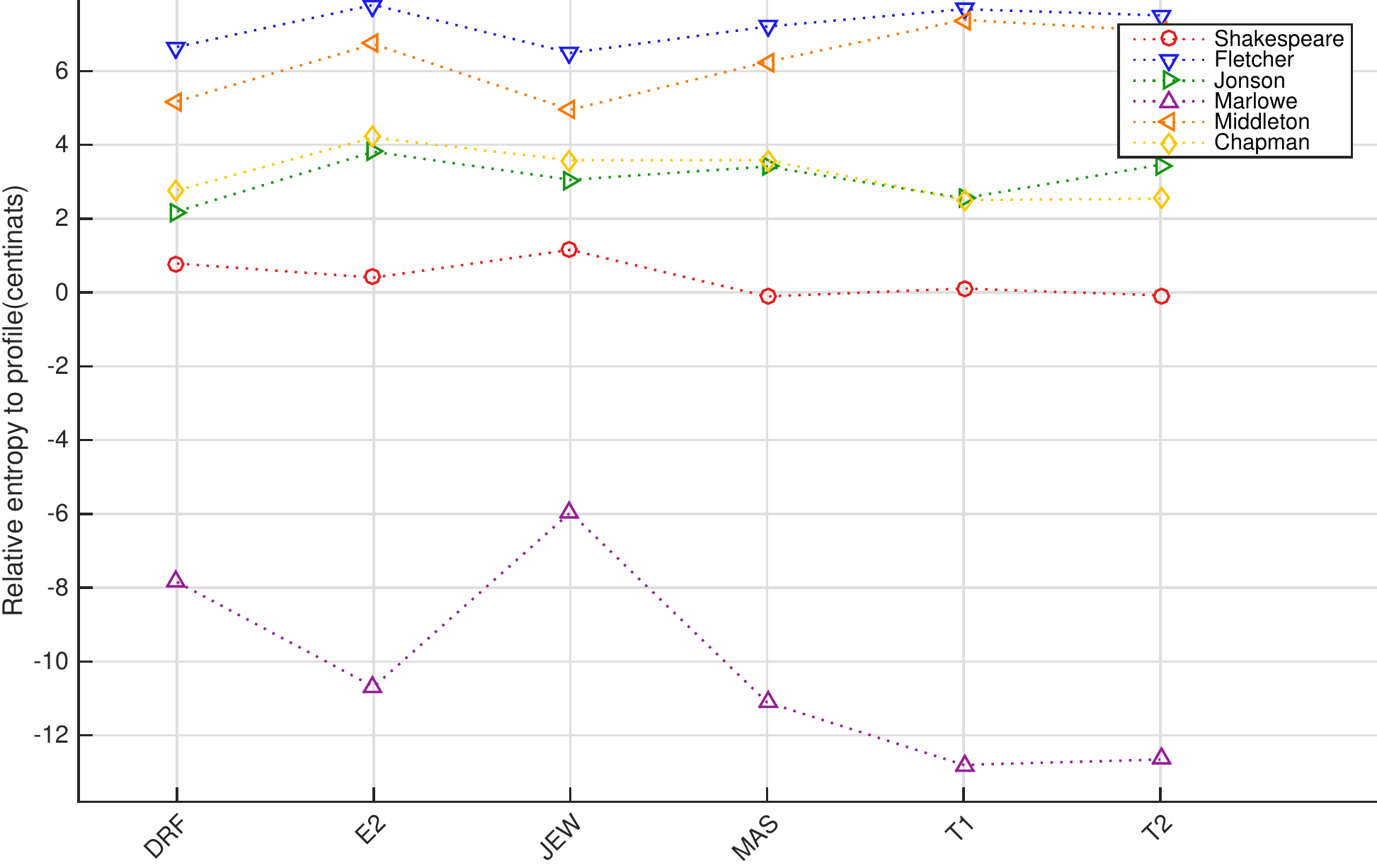}
\caption{Attribution of Marlowe plays. Our method correctly attributes all 6 known-to-be Marlowe plays in Table \ref{table_profile_texts} by a large margin.}
\label{fig_marlowe}
\end{figure}

In Fig. \ref{fig_marlowe}, we present the present method's attribution of 6 plays known to have been written by Christopher Marlowe. Our method achieves an accuracy of 100\% in attributing Marlowe's solo works. Observe that, in the case of sole-authorship plays, each is attributed to Marlowe by a substantial margin  of between $7cn$ and $13cn$. The large negative values (between $-6cn$ and $-13cn$) for the relative entropy between these plays and Marlowe's profile show that they are much closer in style to his profile than they are to the profile of an average playwright. This difference may be due in part to the fact that Marlowe's plays were written at least a decade before most of the other authors considered, thus possibly indicating a shift in writing style during the one or two decades that separate Marlowe from the rest.

\begin{figure*}
\centering
\includegraphics[height=.3\textheight, width = 0.82\textwidth]{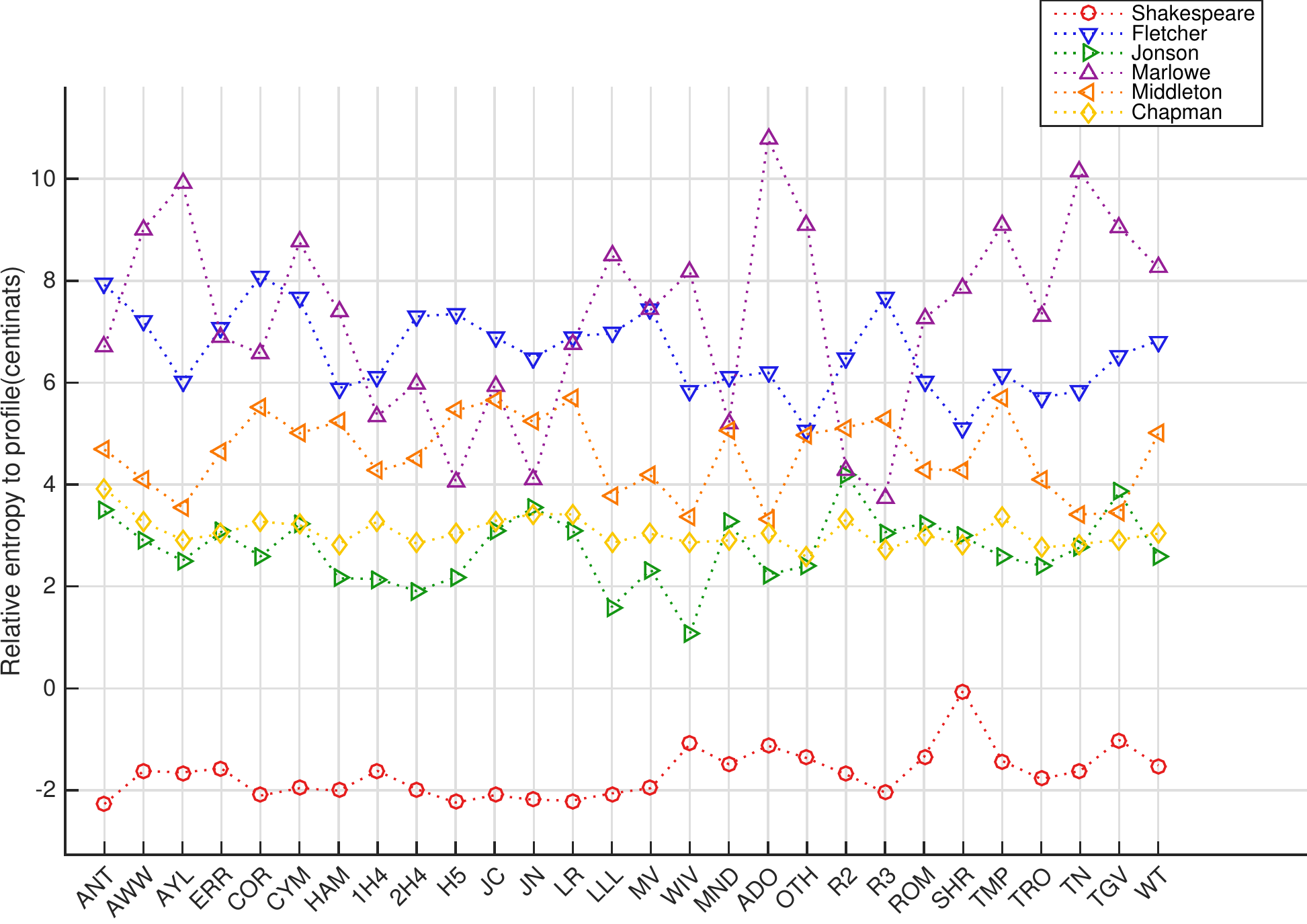}
\caption{Attribution of Shakespeare plays. Our method correctly attributes the 28 known-to-be Shakespeare plays in Table \ref{table_profile_texts}. The distance between Marlowe's profile and each play is smallest for Shakespeare's history plays, suggesting an impact of genre in attribution.}
\label{fig_shakespeare}
\end{figure*}

In Fig. \ref{fig_shakespeare} we show the attribution of 28 plays generally accepted to be written solely by William Shakespeare, and they are correctly attributed to him by our method. It is interesting to observe, however, an exceptional situation in the case of Marlowe. Marlowe's profile is generally very dissimilar from Shakespeare's in Table \ref{table_profile_comparison} and, consequently, he ranks poorly in the attribution of most of Shakespeare's plays, being consistently near the top of Fig. \ref{fig_shakespeare}. However, the relative entropy between Marlowe's profile and the plays \emph{Henry V}, \emph{King John}, \emph{Richard II}, and \emph{Richard III} is around $+4cn$, an uncharacteristically small value compared to the rest of Shakespeare's canon. These four works are all history plays, a genre in which Marlowe wrote \emph{Edward II} and \emph{Massacre at Paris}, comprising a third of his profile. This suggests a potential for genre to confound attributions of authorship, although it may well be a problem confined to the particular genre of history plays (Arefin et al., 2014; Taylor and Loughnane, 2017, pg. 435-6).
\begin{figure*}
\centering
\includegraphics[height=.3\textheight, width =0.82\textwidth]{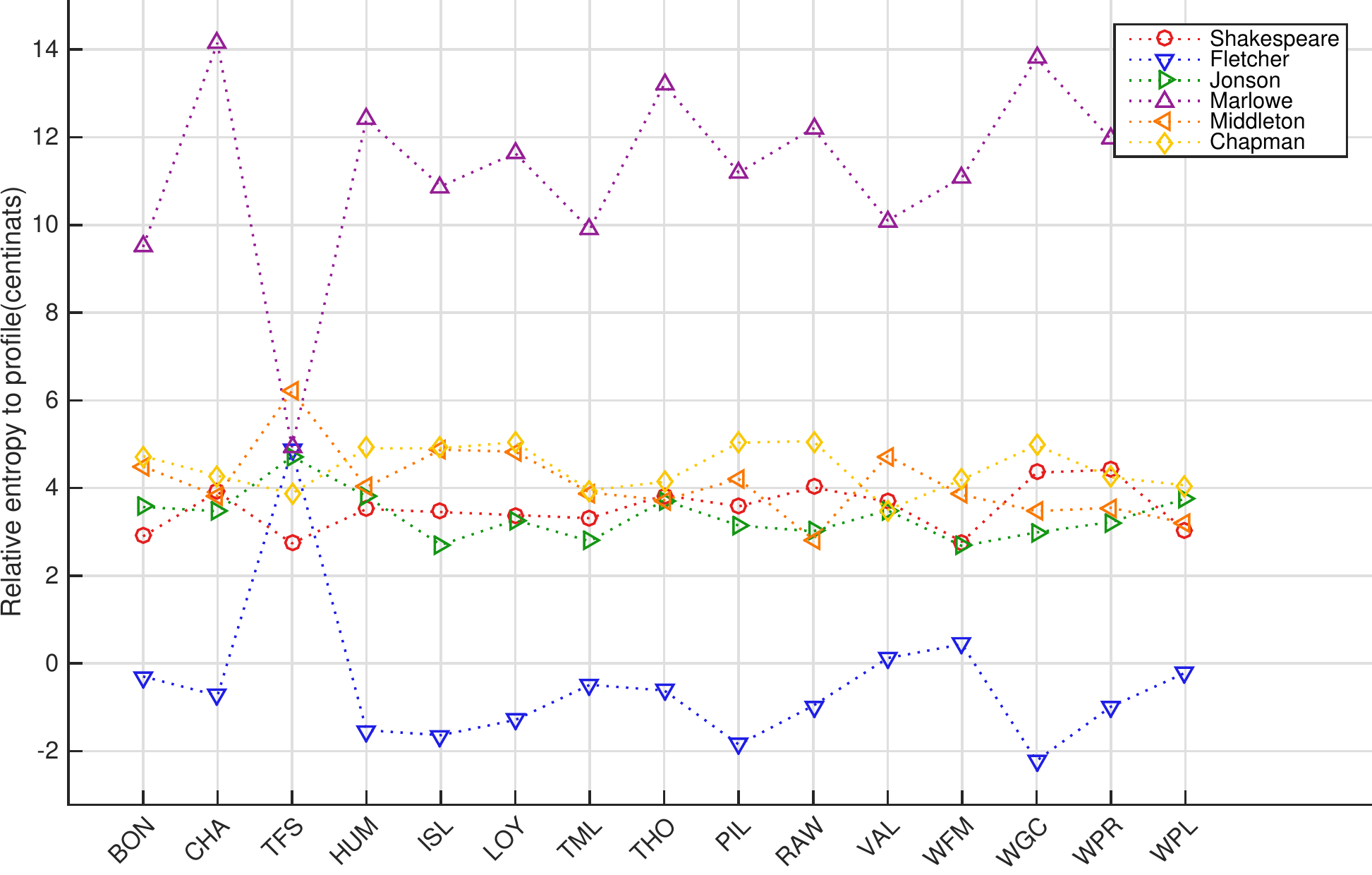}
\caption{Attribution of Fletcher plays. Of the 15 known-to-be Fletcher plays in Table \ref{table_profile_texts}, our method misattributes only \emph{The Faithful Shepherdess}.} 
\label{fig_fletcher_solo}
\end{figure*}

In Fig. \ref{fig_fletcher_solo}, our method attributes the 15 plays of John Fletcher listed in Table \ref{table_profile_texts}. Only the play \emph{The Faithful Shepherdess} is misattributed by our method, with Fletcher notably ranked behind Shakespeare, Chapman, and Jonson. This case is unusual in that the relative entropy between the play and Fletcher's profile is around $+5cn$, with all other Fletcher plays obtaining scores less than $+1cn$.
This finding is consistent with that of Cyrus Hoy in his comprehensive study of the authorship of plays attributed to Fletcher \citep{hoy1956shares}. Hoy concluded that although \emph{The Faithful Shepherdess} is undoubtedly Fletcher's play,``linguistically at least it has nothing in common with any other of his unaided works" because he wrote it in the archaic style of pastoral poetry using forms such as 'hath' and 'doth', which he ``seldom or never uses in his other unaided plays, while all the most distinguishing of his colloquial forms are either completely absent, or present in only a negligible degree" \citep[p. 142]{hoy1956shares}. Thus, writes Hoy,``Nothing could be more misleading than to regard the language of \emph{The Faithful Shepherdess} as typically Fletcherian". It is salutary to note that when a writer departs markedly from his usual style he can confound studies that attribute authorship by analysis of style, but nonetheless comforting that our method corroborates a judgment made long ago by the world's leading expert on this author's style, using methods quite different from ours.

\subsection{Comparison with Existing Methods}\label{sec_summary}

\begin{table*}
\caption{Accuracies of various attribution methods on full plays between six candidate authors.}
\label{table_method_comparison}
\centering
\renewcommand{\arraystretch}{1.3}
\input{table_method_comparison.tex}
\end{table*}

In total, our method attributes correctly 87 out of the 94 single-authored and elsewhere reliably attributed plays in Table \ref{table_profile_texts}, yielding an accuracy of $92.6\%$. Furthermore, if we consider only the attributions in which we have the greatest confidence, i.e. among authors that are more than $5 cn$ apart in Table~\ref{table_profile_comparison}, then we fail only in 4, yielding an accuracy of $95.7\%$. To compare the power of our method in attributing plays from this era with the power of other methods commonly used in Shakespeare attributional studies, we run the same validation tests on the 94 plays using two common frequency-based methods. The first, known as Burrows's Delta method \citep{burrows2002delta}, involves measuring word frequency vectors for each play. The frequency vectors are normalized to Z-scores and compared with one another using a distance metric. The mean distance of the Z-score vector of an unknown play to the Z-score vectors of the plays of a candidate author determines the distance of a play to the candidate author. Various metrics are used, the most common being Manhattan distance, Euclidean distance, and cosine similarity. The Manhattan distance computes the sum of the absolute value of the difference in each component of the Z-score vectors. Alternatively, the Euclidean distance computes the sum of the squared difference in each component of Z-score vectors. The cosine similarity, on the other hand, computes the similarity as the cosine of the angle between the two Z-score vectors, which rises from 0 to 1 as the size of the angle between them increases from 0 degrees (when the vectors lie on top of one another, showing maximum similarity) to 90 degrees (when the vectors are orthogonal, showing maximum dissimilarity).

We additionally compare the accuracy of the WAN method against the PCA-based method used in \citep{craig2009shakespeare}. Word frequency vectors are again constructed, but now reduced to principal component (PC) score vectors. This method reduces the dimensionality of the word frequency vectors to contain only the components with highest variation, known as principal components. The play is attributed to the author whose PC score vectors have the smallest mean distance to the target play's PC score vector. In this case, the method can be varied by using more or fewer principal components.

In Table \ref{table_method_comparison} we compare the accuracy of the WAN method against the Delta and PCA methods when attributing plays among the six authors. For each method, we choose the number of function words that maximized attribution accuracy. Observe that the WAN method achieves an accuracy of $92.6\%$, outperforming five variations of the aforementioned methods. The closest competing strategy is the Delta method with Manhattan distance, which achieved an accuracy of $91.3\%$. All other methods achieve accuracies lower than $82\%$. We stress the high classification power of the WAN method for plays of sole authorship relative to other popular methods.

\section{Collaborations}\label{sec_collaborations}

In cases of multiple authors contributing to a single play, we can show how our method is able to detect one or more of the authors present. We illustrate this ability in two ways: (i) by attributing collaborative plays to profiles built from other collaborations, and (ii) by attributing collaborative plays to profiles built from sole-authored plays for each contributing author.

\subsection{John Fletcher and Collaborators}\label{sec_fletcher}

John Fletcher wrote numerous plays both by himself and with collaborators, making him a suitable case study for how our method copes with co-authorship. In addition to the six profiles built for sole-authored plays in the previous section, we now include two profiles built from plays written by Fletcher in collaboration with his two most frequent co-authors: Francis Beaumont and Phillip Massinger; see Table~\ref{table_profile_texts}.

\begin{figure}
\centering
\includegraphics[height=.3\textheight, width=0.8\textwidth]{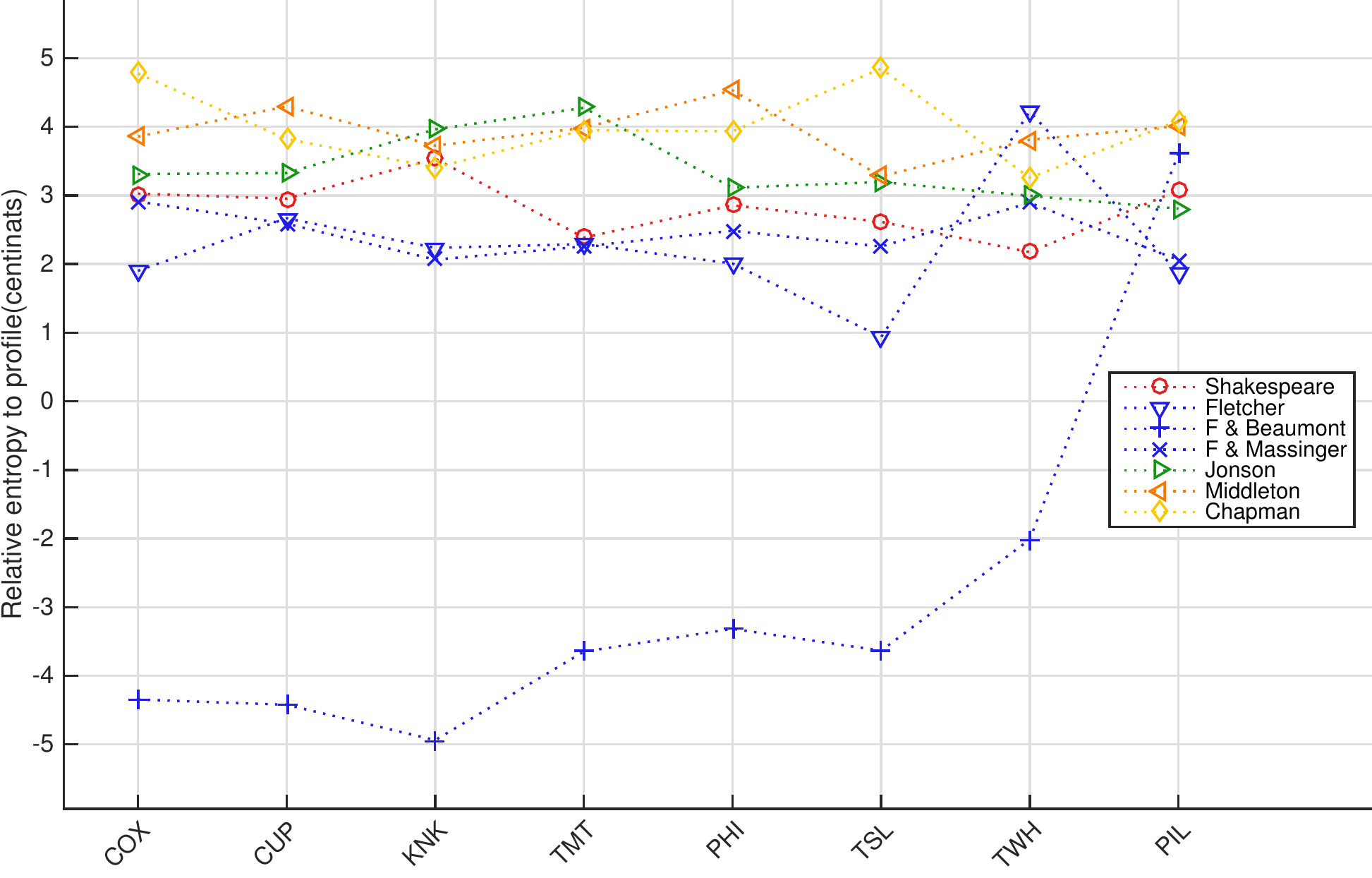}
\caption{Attribution of Fletcher and Beaumont plays. A single play, \emph{Love's Pilgrimage}, is here wrongly attributed by our method to the solo Fletcher canon instead of the Fletcher \& Beaumont canon where it belongs.}
\label{fig_fletcher_collab1}
\end{figure}

\begin{figure}
\centering
\includegraphics[height=.3\textheight, width=0.8\textwidth]{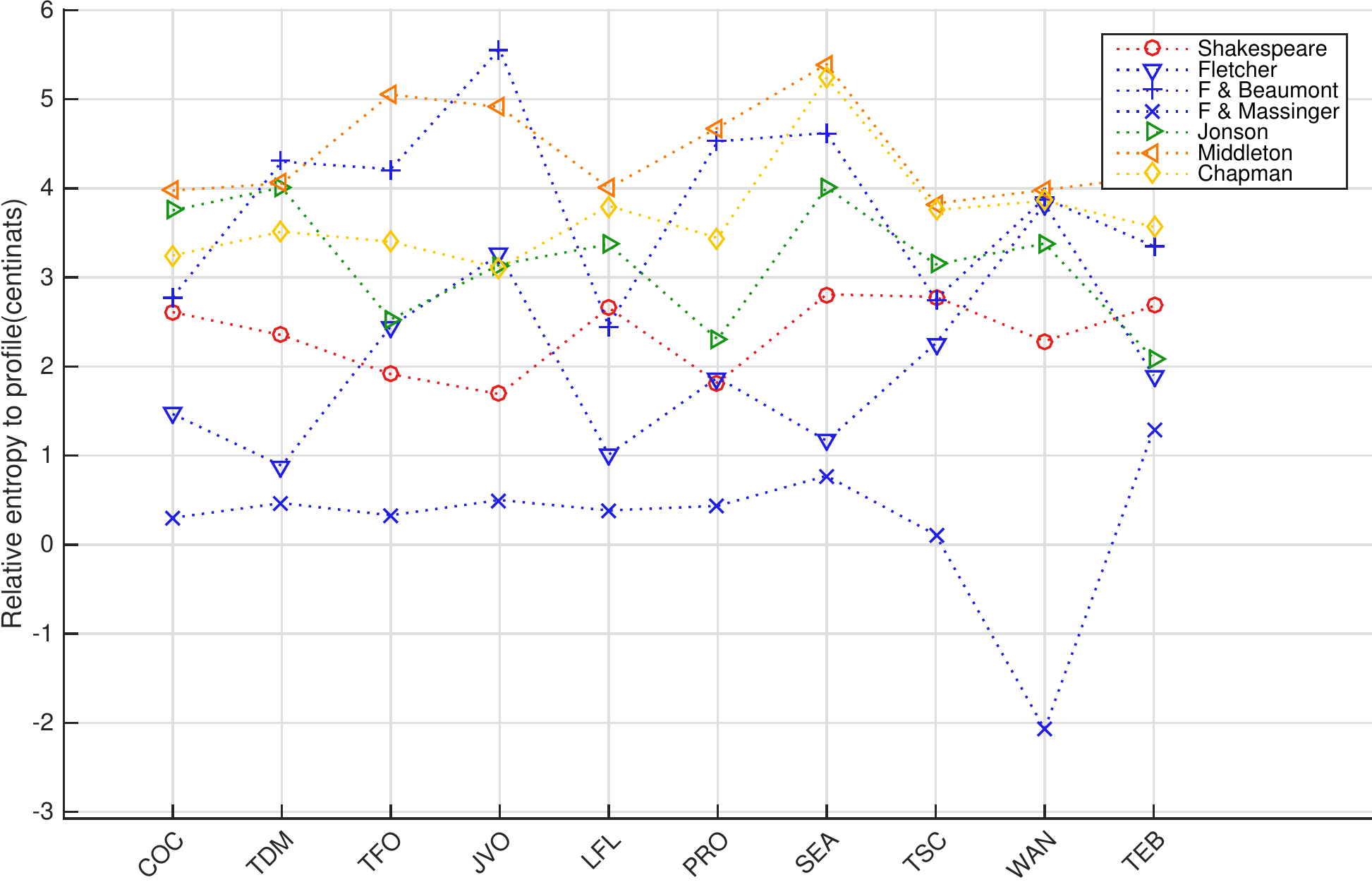}
\caption{Attribution of Fletcher and Massinger plays. All plays are correctly attributed to the Fletcher \& Massinger canon by our method, with the solo Fletcher canon often ranked second.}
\label{fig_fletcher_collab2}
\end{figure}

The attribution of Fletcher's collaborative works with Beaumont is shown in Fig. \ref{fig_fletcher_collab1} while the attribution of Fletcher's collaborative works with Massinger is shown in Fig. \ref{fig_fletcher_collab2}. In both figures we omit the marker corresponding to Marlowe since he is poorly ranked for every play. This is consistent with Fletcher and Marlowe having the most dissimilar writing styles; see Table \ref{table_profile_comparison}.

In Fig. \ref{fig_fletcher_collab1}, seven of the eight Fletcher and Beaumont plays are correctly attributed by our method to the Fletcher and Beaumont canon. A single mistake occurs for \emph{Love's Pilgrimage}, although even for that play, the solo Fletcher profile and Fletcher and Massinger profile are ranked nearest and second nearest, respectively. Additionally, ten Fletcher and Massinger plays are correctly attributed by our method to the Fletcher and Massinger canon, as can be seen in Fig. \ref{fig_fletcher_collab2}. Observe that, in many cases, the solo Fletcher profile is ranked second behind the correct collaborative profile. These results demonstrate a case in which the WAN method is not only able to distinguish between single author in plays, but is able to distinguish between an author's collaborations with two different authors. While this is not a comprehensive study of the method's discriminative power, it suggests that multiple authorial styles can be encoded in the WAN structure simultaneously. 

\subsection{Other Collaborations}

\begin{figure*}
\centering
\includegraphics[height=.3\textheight, width=0.82\textwidth]{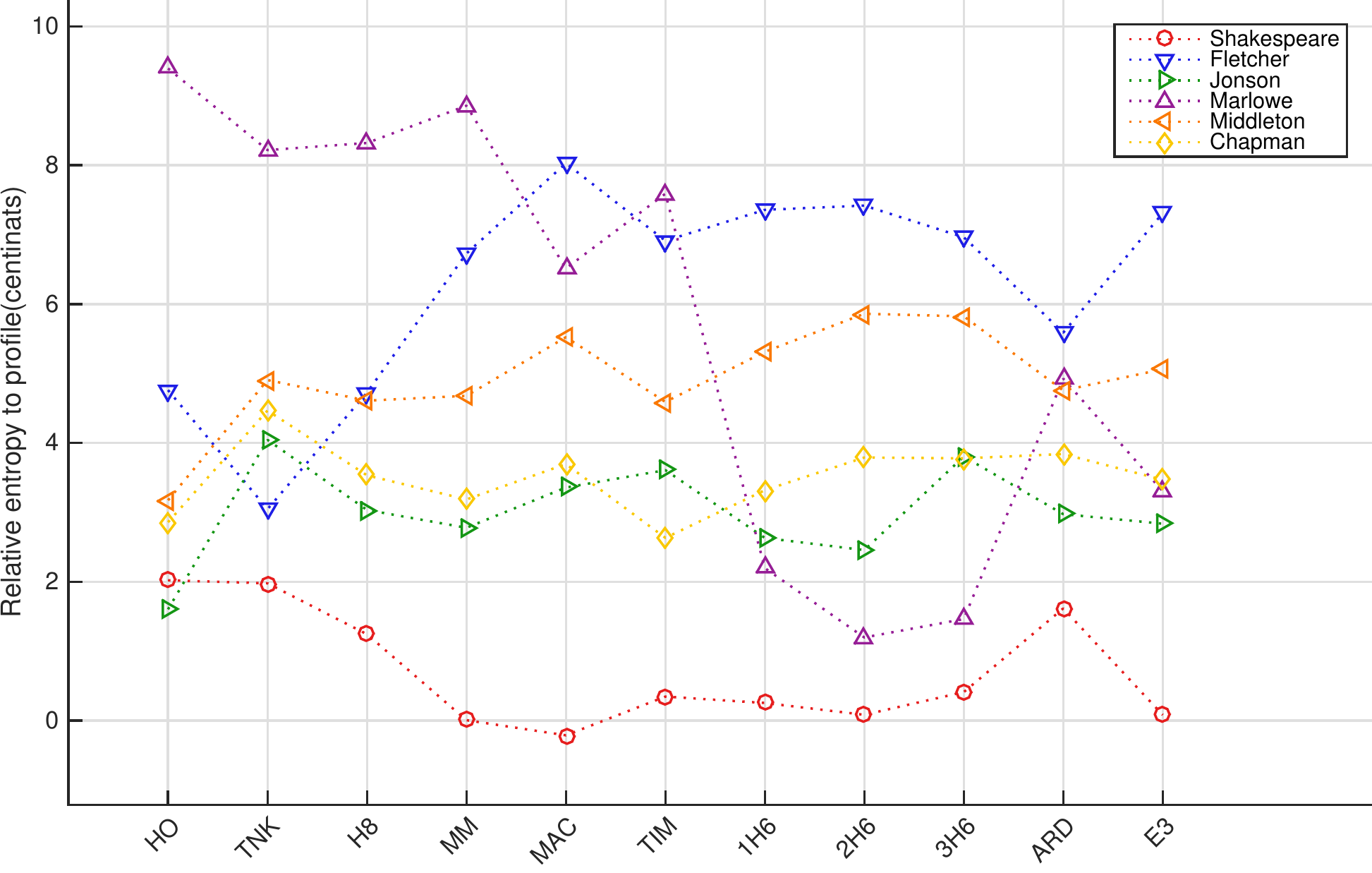}
\caption{Attribution by our method of collaborative plays listed in Table \ref{table_att_texts}. All plays here are attributed to one of the commonly proposed contributing authors, with the other contributing author often ranking second or third.}
\label{fig_collab}
\end{figure*}

A shortcoming of the attribution method used to attribute the collaborations of Fletcher with Beaumont and Massinger is that it requires multiple collaborations between two authors so that a reliable profile of each collaboration can be built. We also examine, therefore, the case in which we attribute a collaborative play using only profiles built from single authors. In Table \ref{table_att_texts} in the Appendix we list a set of plays with either undisputed or speculated collaboration between multiple authors previously profiled. These plays are attributed by our method by observing which of the six original author profiles are closest to each play's WAN, with the results shown in Fig. \ref{fig_collab}.

First, we attribute the play \emph{Eastward Ho}, generally accepted as a collaboration between Jonson and Chapman and a third author, John Marston, whom we have not profiled. By our method, Jonson and Chapman are indeed ranked first and third, respectively. We also attribute two well-known collaborations between Shakespeare and Fletcher, namely \emph{Henry VIII} and \emph{The Two Noble Kinsmen}. We attribute both to Shakespeare, with Fletcher the second preferred author in the latter. In the case of the former, on the other hand, Fletcher is not well ranked and his contribution is not evident from the attribution of the entire play; we cannot account for this. The attribution of Shakespeare's collaborations with Middleton, on the other hand, do not suggest the presence of both authors. While all three plays, \emph{Measure for Measure}, \emph{Macbeth}, and \emph{Timon of Athens} are correctly attributed to Shakespeare, Middleton is ranked very poorly being the fourth closest candidate in all of them. This is consistent with the accepted idea that Middleton's contribution to the first two plays is minimal, but the most recent study of \emph{Timon of Athens} attributes to him about a third of the lines and we cannot explain his poor showing here, although there is more to be said on this topic below (Jackson, 1979; Taylor and Lavagnino, 2007, pg. 467;  Wells, 2009).

We also perform an attribution of a set of plays often considered to be collaborations between Shakespeare and Marlowe, though with less scholarly consensus than the previous examples. In the full play attribution of \emph{1 Henry VI}, \emph{2 Henry VI}, and \emph{3 Henry VI}, however, our method shows that Shakespeare is the strongest presence with Marlowe ranked second. This is notable because Marlowe is generally ranked very poorly when attributing Shakespeare's other works as shown in Fig. \ref{fig_shakespeare}. While a genre bias towards Marlowe was evident with Shakespeare's other history plays,  the relative closeness to Marlowe's profile in these cases is so striking that this bias cannot be the full explanation: Marlowe's hand is also definitely present here. The WAN method is used to further examine Marlowe's role in this trilogy in \citep{segarra2016attributing}. The attribution of the anonymously published \emph{Arden of Faversham} and \emph{Edward III} both support theories of Shakespearean authorship \citep{vickers2002shakespeare}. 

We continue by providing a more detailed study of the collaboration in these works in Section \ref{sec_collab_intraplay} by breaking each play into smaller components.

\section{Collaborations -- Intraplay Analysis}\label{sec_collab_intraplay}

\begin{table*}
\caption{Accuracies of various attribution methods on acts and scenes among eight and two authors, respectively.}
\label{table_method_comparison_scene}
\centering
\renewcommand{\arraystretch}{1.3}
\input{table_method_comparison_scene.tex}
\end{table*}

We examine the authorship of collaborative plays through the attribution of its individual acts and scenes. In Section \ref{sec_collaborations} we attempted to detect collaborations in full plays by looking at the top candidate authors. This does not, however, suggest any particular breakdown of which sections of the text were contributed by which author. To pursue that topic, we divide plays into acts and scenes and attempt to attribute these individually to gain deeper insight as to how the play was written. We also see cases where we can detect collaboration through intraplay analysis where we could not when attributing the full text.

In the following sections we attribute plays of known or suggested collaboration between eight candidate authors: the six previously introduced plus Robert Greene and George Peele. These two additional authors were not included previously because their canons are small but are included here because they have been suggested as candidates for collaboration for some of the plays that we are considering. The plays used to build Greene's and Peele's author profiles are listed in Table~\ref{table_profile_texts}. 

We first re-train the WAN networks due to the fact that, counterintuitively, smaller WANs may increase the attribution accuracy when working with shorter texts. This is because shorter texts are less likely to contain uncommon function words. As a result, larger networks that contain these uncommon function words are more prone to overfit to features of specific texts rather than the features of the broader authorial style. We must here also point out that, due to the short length of scenes and the resulting diminution in our method's power to discriminate between authors, we only seek to distinguish the most plausible of the two most commonly cited candidates when working with scenes, rather than selecting among all eight candidates as we do when working with acts. To re-train, we divide each text in Table \ref{table_profile_texts} into acts and scenes and find the network size that correctly attributes the greatest number of these units to its known author. Note that, when attributing a particular act or scene of a play, the \emph{entire} play is removed from the corresponding author profile to avoid bias.

Our training procedure finds that using the 76 and 55 most common functions words optimally achieve accuracies of $93.4\%$ and $91.5\%$ for attributing acts and scenes, respectively, of the plays in Table~\ref{table_profile_texts}. The words used in these reduced-size networks are listed in Table \ref{table_function_words_all}. As a point of comparison, we include in Table \ref{table_method_comparison_scene} the attribution accuracies achieved by the Delta and PCA-based methods on acts and scenes. In this case, the WAN method outperforms all the other methods by significant margins for both acts and scenes. The largest accuracies achieved by the alternative methods are $74.3\%$ and $71.5\%$ for acts and scenes, respectively. We stress the high classification power of the WAN method relative to the other attribution schemes when attributing individual acts and scenes.

\begin{figure}[t]
\centering
\includegraphics[height=.25\textheight, width=.5\textwidth, keepaspectratio]{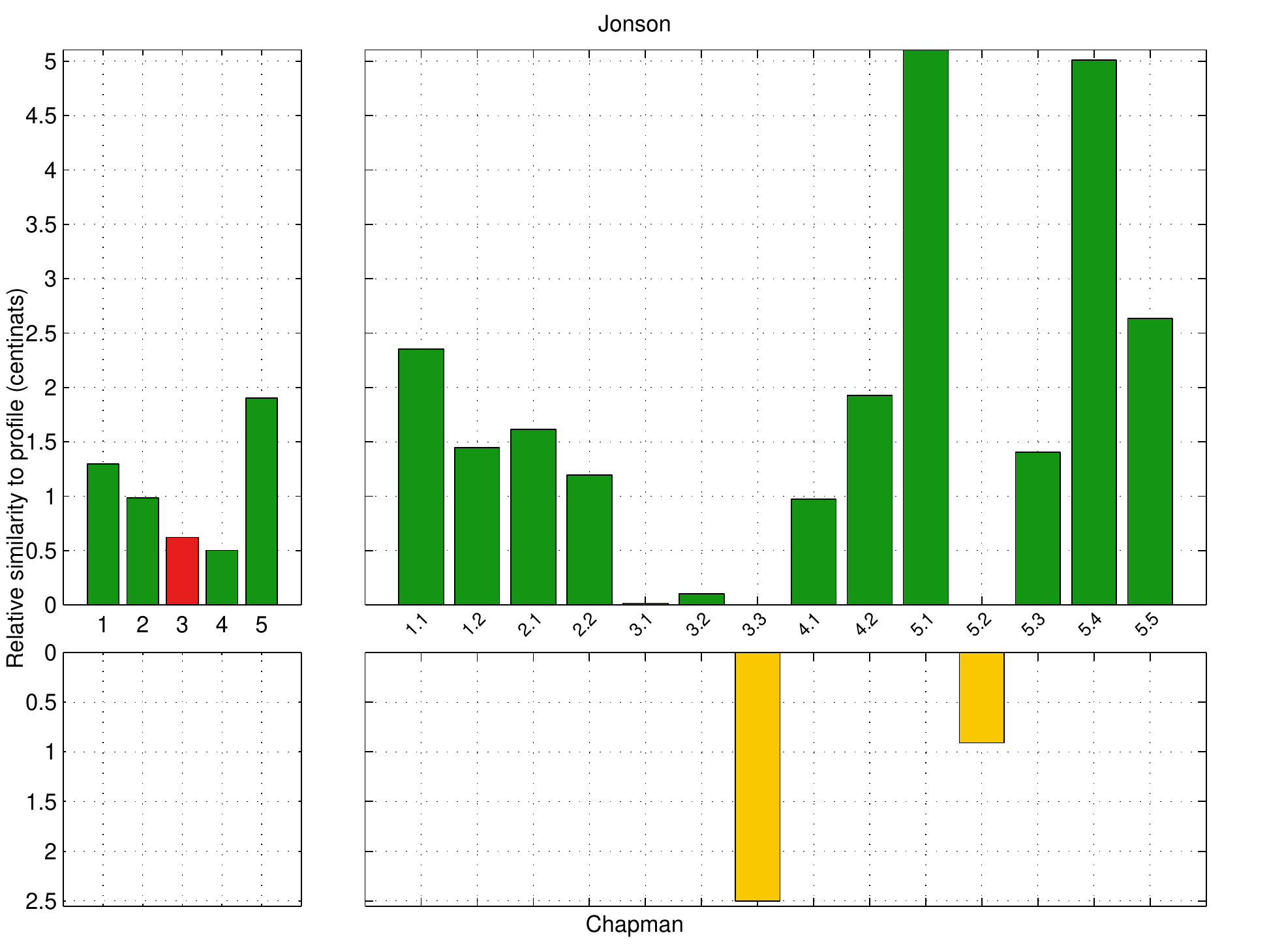}
\caption{Attribution of acts and scenes of \emph{Eastward Ho}. Act 3 is assigned to Shakespeare (red) over both Jonson (green) and Chapman (yellow). This attribution is an exception to the typical case, due to the fact that a third author (Shakespeare) is ranked first for one of the attributions; we nevertheless perform the scene-wise comparison for just Jonson and Chapman, because they are the widely accepted co-authors.}
\label{fig_eastward_ho_attribution}
\end{figure}

\begin{figure}[b]
\centering
\includegraphics[height=.25\textheight, width=.5\textwidth, keepaspectratio]{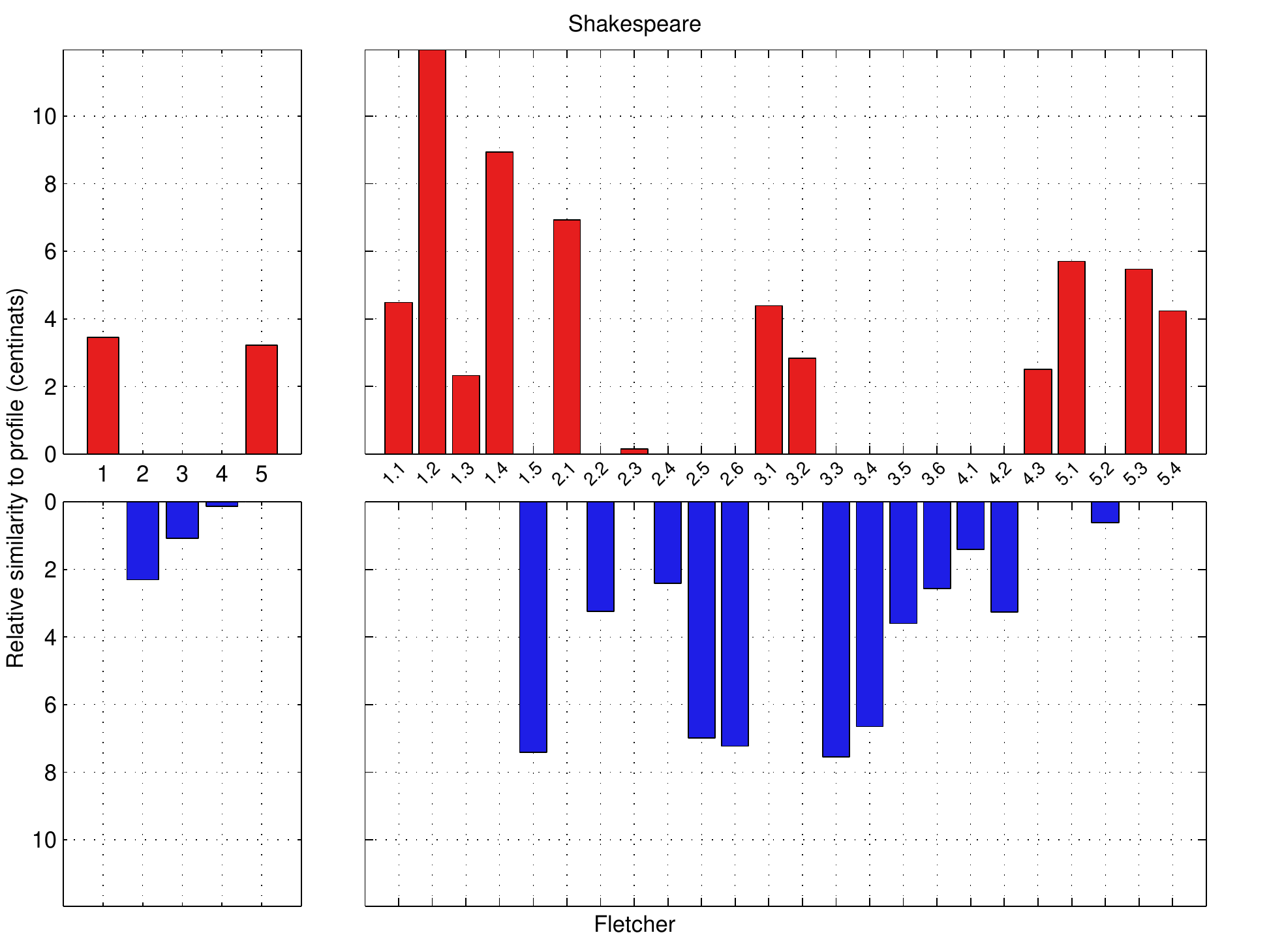}
\caption{Attribution of acts and scenes of \emph{Two Noble Kinsmen} between Shakespeare (red) and Fletcher (blue).}
\label{fig_two_noble_kinsmen_attribution}
\end{figure}

In the following subsections, the figures display the difference in relative entropy for acts and scenes when comparing the two top candidate authors, reflected by both the color of the bars and the titles above and below the plot. The longer the bar in a particular direction, the larger the difference between the entropies of the two top candidate authors. For example, in Fig. \ref{fig_two_noble_kinsmen_attribution},  red bars extending upwards indicate an attribution to Shakespeare while blue bars extending downwards indicate an attribution to Fletcher. The attribution of acts is performed between eight candidate authors, though we only plot the distances to the two most highly ranked by our method for ease of viewing. In the attribution of scenes, on the other hand, we consider only the two authors most often cited as candidates. In many cases, the acts and scenes are attributed among the same pair of authors. Cases in which an act is attributed to a third author--because our method proposes a candidate not previously considered by most other investigators--are marked in the figure captions. We note that, after Fig. \ref{fig_eastward_ho_attribution}, all future figures appear in the appendix at the end of the document.

\subsection{Jonson and Chapman}\label{sec_jonson_chapman}

We attribute both the individual acts and scenes of the single known collaboration between Jonson and Chapman, \emph{Eastward Ho}, which also includes contributions from a third author, John Marston. Fig. \ref{fig_eastward_ho_attribution} displays the results of the act (left) and scene (right) attribution. In the eight author comparison, every act is assigned to Jonson, with the exception of Act 3 assigned to Shakespeare. Chapman is ranked either third or forth in all acts except Act 3 in which he is ranked second. These results are similar to the full play attribution from Fig. \ref{fig_collab}, in which Jonson was the top ranked author and Chapman ranked third. While these results on their own do not highlight Chapman's specific contribution, a look at the scene attribution between just Jonson and Chapman illuminates some of Chapman's possible contributions. Most of the play is still assigned to Jonson, however Chapman is seen as a more likely candidate in Scenes 3.3 and 5.2 whereas the attribution of Scenes 3.1 and 3.2 is too close to make any conclusion. While there is not a scholarly consensus on the scene breakdown, many attribute Marston to Act 1, Chapman to Act 2 and 3, and Jonson to Act 5 \citep{logan1977new}. Most scholars agree in particular about Scene 3.3 being written by Chapman \citep{van1979eastward}. Our results support the notion that Chapman did not write Act 1 and Jonson wrote Act 5. We also provide further evidence that Chapman wrote 3.3, as it is, in our analysis, the single scene that is assigned to Chapman by a margin larger than $2cn$. We also, however, find more evidence of Jonson than Chapman contributing Acts 2 and 4.

\subsection{Shakespeare and Fletcher}\label{sec_shakespeare_fletcher}

\begin{figure}[t]
\centering
\includegraphics[height=.25\textheight, width=.5\textwidth, keepaspectratio]{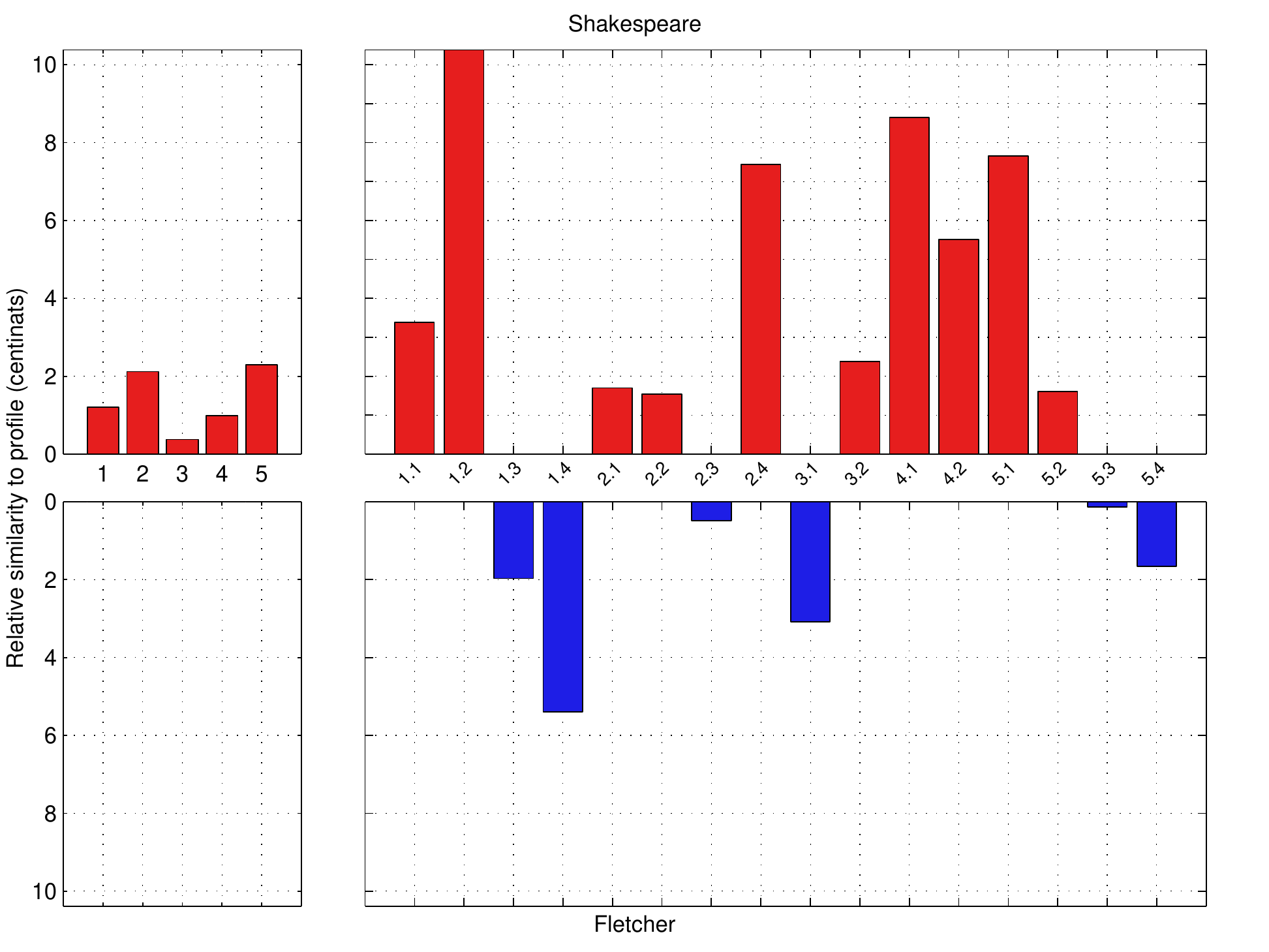}
\caption{Attribution of acts and scenes of \emph{Henry VIII} between Shakespeare (red) and Fletcher (blue).}
\label{fig_henry_viii_attribution}
\end{figure}

\begin{figure}[b]
\centering
\includegraphics[height=.25\textheight, width=.5\textwidth, keepaspectratio]{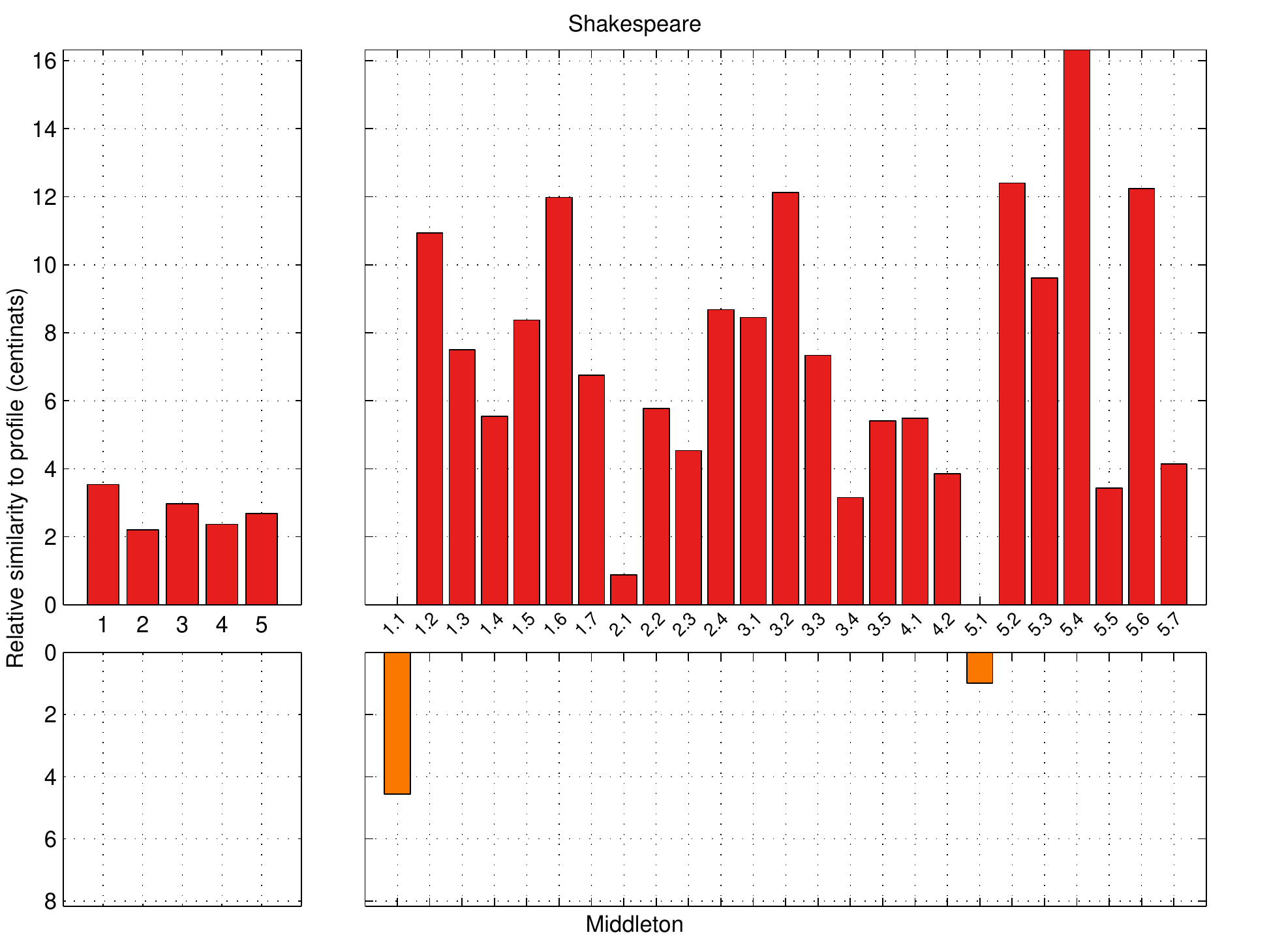}
\caption{Attribution of acts and scenes of \emph{Macbeth} between Shakespeare (red) and Middleton (orange).}
\label{fig_macbeth_attribution}
\end{figure}

In Fig. \ref{fig_two_noble_kinsmen_attribution} we show the attribution of individual acts and scenes of \emph{Two Noble Kinsmen}, a known collaboration between Shakespeare and Fletcher. Whereas in Fig. \ref{fig_collab} the play as a whole is assigned to Shakespeare with Fletcher as the second best candidate, here Acts 1 and 5 are assigned to Shakespeare while Acts 2 and 3 are assigned to Fletcher. Act 4 is assigned to Fletcher with Shakespeare and Jonson close behind. Recall that, in Fig. \ref{fig_two_noble_kinsmen_attribution} distances to only the two closest candidates are shown for ease of viewing. A closer look into the scene breakdown, where we consider only Shakespeare and Fletcher as candidates, reveals more specific assignments. Shakespeare is assigned Scenes 1.1, 1.2, 1.3, 1.4, 2.1, 3.1, 3.2, 4.3, 5.1, 5.3, and 5.4; Fletcher is assigned to Scenes 1.5, 2.2, 2.4, 2.5, 2.6, 3.3, 3.4, 3.5, 3.6, 4.1, and 4.2; and close ties in Scenes 2.3 and 5.2. The scene breakdown we propose largely supports the one given by Hallet Smith in \emph{The Riverside Shakespeare} \citep{shakespeare1974riverside}.

\begin{figure}[t]
\centering
\includegraphics[height=.25\textheight, width=.5\textwidth, keepaspectratio]{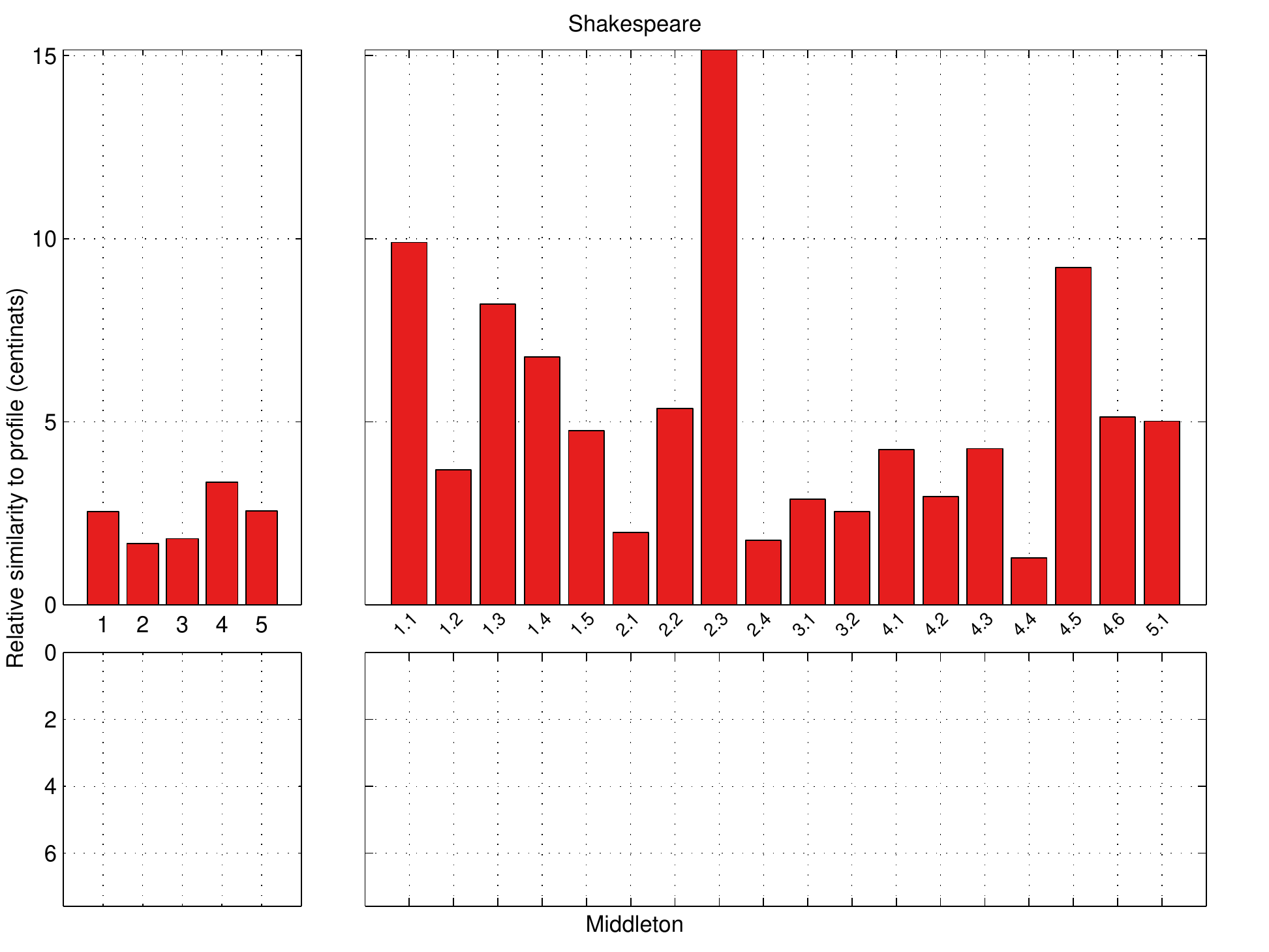}
\caption{Attribution of acts and scenes of \emph{Measure for Measure} between Shakespeare (red) and Middleton (orange).}
\label{fig_measure_for_measure_attribution}
\end{figure}

The act and scene analysis of Shakespeare and Fletcher's other collaboration--\emph{Henry VIII}--is displayed in Fig. \ref{fig_henry_viii_attribution}. Recall that, when attributing the full play, Shakespeare was the top candidate while Fletcher was in fact ranked fourth, thus revealing no substantial evidence of collaboration; see Fig. \ref{fig_collab}. We see similar results in Fig. \ref{fig_henry_viii_attribution}, in which Shakespeare, in an eight-author act-wise comparison, is assigned every act. Fletcher, again, is ranked poorly in every act. A scene-by-scene analysis involving just Shakespeare and Fletcher however, does reveal Fletcher to be a stronger candidate than Shakespeare in several individual scenes. In fact, the scene breakdown we observe---in which Shakespeare is assigned scenes 1.1, 1.2, 2.1, 2.2, 2.4, 3.2, 4.1, 4.2, 5.1, and 5.2; Fletcher is assigned scenes 1.3, 1.4, 3.1, and 5.4; and 2.3 and 5.3 are too close to call---is aligned to that proposed by Cyrus Hoy \citep{hoy1960shares} and currently accepted by many scholars. The primary area of disparity between the breakdown we propose and the one given by Hoy is the authorship of Act 4. While Hoy assigns Act 4 to Fletcher,  we find that there is greater evidence that Shakespeare contributed this section. Both scenes in Act 4 are attributed by our method to Shakespeare by a significant margin of at least $5cn$. Another point of contention is that we assign 2.3---given to Shakespeare by Hoy---to Fletcher by a small margin. 

The attribution of \emph{Henry VIII} shows that we may detect collaboration at the level of scenes that may be undetectable when looking at entire plays or acts. In this play, there are several individual scenes that we attribute to Shakespeare by a margin as wide as $7cn$, such as Scenes 1.2, 2.4, 4.1, and 5.1. When an act contains scenes by different authors and some of the scenes have such high scores, this may tend to bias the attribution of complete acts, while the scene-by-scene analysis provides a clearer perspective.

\subsection{Shakespeare and Middleton}\label{sec_shakespeare_middleton}

\begin{figure}[t]
\centering
\includegraphics[height=.25\textheight, width=.5\textwidth, keepaspectratio]{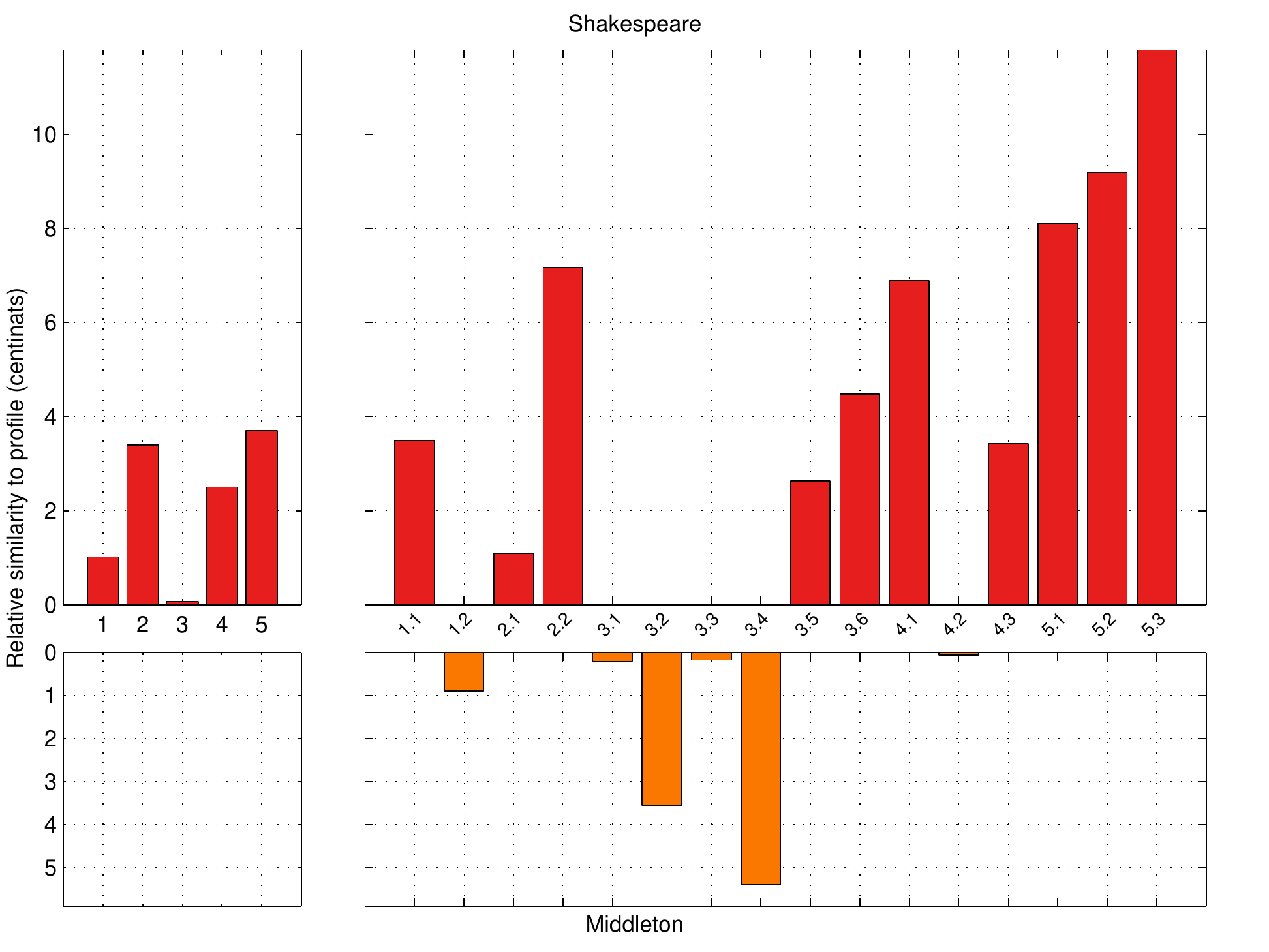}
\caption{Attribution of acts and scenes of \emph{Timon of Athens} between Shakespeare (red) and Middleton (orange).}
\label{fig_timon_athens_attribution}
\end{figure}

We analyze in Figs. \ref{fig_macbeth_attribution}-\ref{fig_timon_athens_attribution} Middleton's contributions to Shakespeare's plays, \emph{Macbeth}, \emph{Measure for Measure}, and \emph{Timon of Athens}. The attribution of the full plays in Fig. \ref{fig_collab} did not suggest that Middleton made any significant contribution to any of these plays. The intraplay analysis of \emph{Macbeth} at the level of acts and scenes, shown in Fig. \ref{fig_macbeth_attribution}, supports this conclusion. A total of two scenes are assigned to Middleton over Shakespeare, namely Scenes 1.1 and 5.1. Scene 5.1 is attributed to Middleton by only a small margin of $1cn$ while Scene 1.1 is assigned by a more substantial margin of over $4cn$. Scholars have often flagged Scenes 1.2, 3.5, and 4.1 as scenes revised or contributed by Middleton \citep{wells2009shakespeare}, although we do not find evidence of this in our analysis.

The case of \emph{Measure for Measure} favors Shakespeare's sole authorship even more; both the act and scene analysis displayed in Fig. \ref{fig_measure_for_measure_attribution} find Shakespeare to be the sole author of the play. If Middleton revised the original play as proposed by scholars \citep{wells2009shakespeare, taylor1993shakespeare}, we do not find evidence that it comprised substantial fresh writing.

Of the three plays, we find that Middleton's contribution was likely largest in \emph{Timon of Athens}. While all five acts are attributed by our method to Shakespeare, in Act 3 it is by a margin of less than $1cn$ from Middleton; see Fig. \ref{fig_timon_athens_attribution}. This is even more evident in the scene analysis. Middleton is a stronger candidate in Scenes 1.2, 3.2, and 3.4, with close ties in Scenes 3.1, 3.3, and 4.2. This assignment supports much of the claim of authorship provided in \citep{vickers2002shakespeare, wells2009shakespeare}, and is broadly consistent with the most thorough analysis that numbers only the scenes, 1 through 19 \citep[pg. 467]{taylor2007thomas}.

\subsection{Shakespeare and Marlowe}\label{sec_shakespeare_marlowe}

Although there are no unanimously agreed upon collaborations between Shakespeare and Marlowe, there exist a number of plays with controversial authorship that have been the subject of scholarly treatment regarding Marlowe's contributions. Of the traditional Shakespeare canon, the three parts of \emph{Henry VI} are the most common points of contention. We analyze the authorship of these plays using the WAN method in depth in \citep{segarra2016attributing}. Here we expand this analysis to include two anonymously published plays that have previously been attributed at least in part to Shakespeare and Marlowe, namely \emph{Arden of Faversham} and \emph{Edward III}.

\begin{figure}[b]
\centering
\includegraphics[height=.25\textheight, width=.5\textwidth, keepaspectratio]{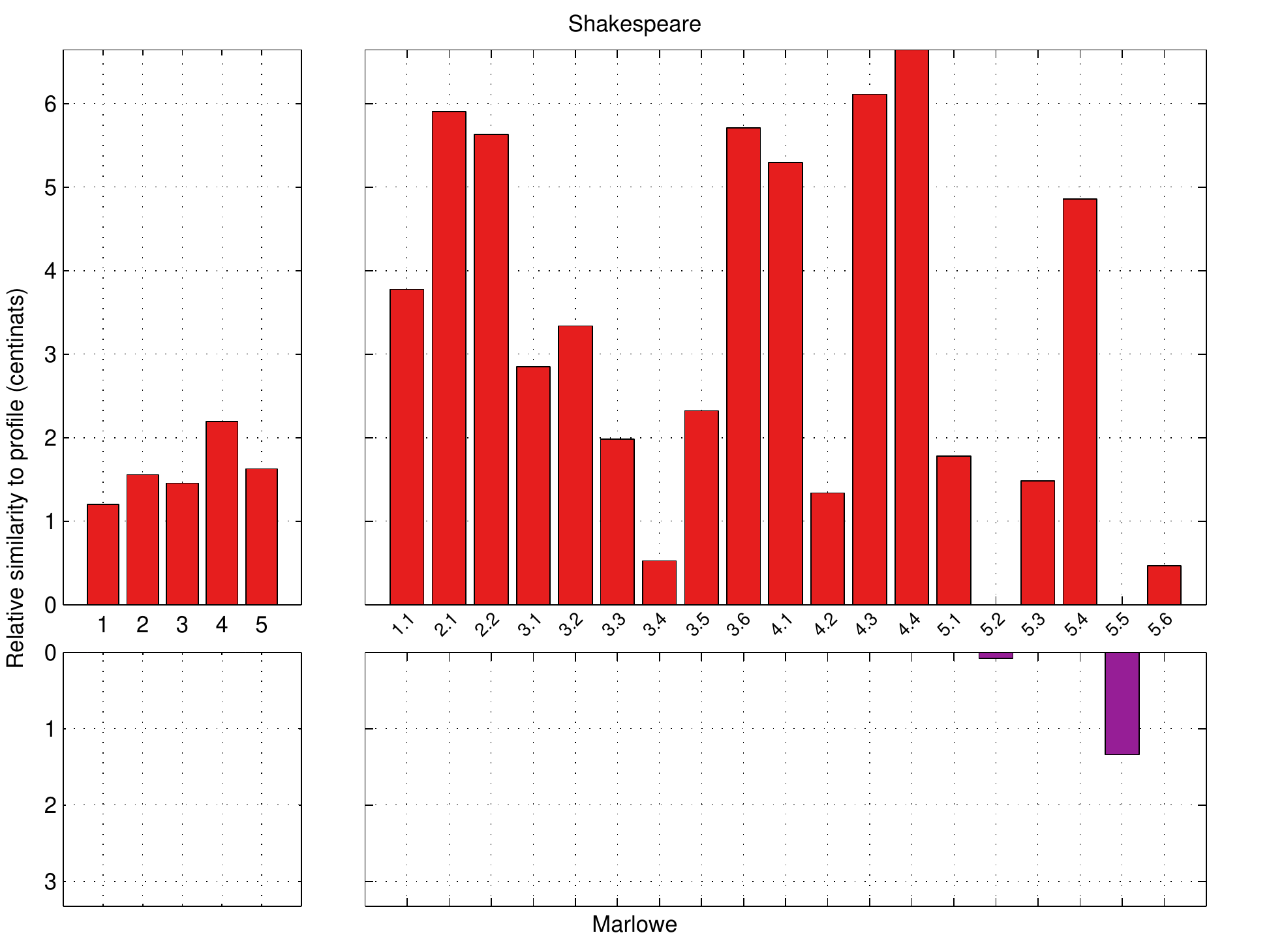}
\caption{Attribution of acts and scenes of \emph{Arden of Faversham} between Shakespeare (red) and Marlowe (purple).}
\label{fig_arden_faversham_attribution}
\end{figure}

We perform in Fig. \ref{fig_arden_faversham_attribution} the intraplay analysis on the play \emph{Arden of Faversham}. Every act is attributed here to Shakespeare ahead of our seven other candidates. Although not shown in the figure, the second preferred candidate in all acts except Act 5 is Jonson, who is not generally thought to have started writing plays until the late 1590s, while \emph{Arden of Faversham} was written between 1587 and 1592 \citep[pg.497-90]{taylor2017canon}. Jonson's immaturity (he was born in 1572) makes him an unlikely candidate unless \emph{Arden of Faversham} was written at the end of its possible date-range. The other commonly considered candidates for authorship are Thomas Kyd and Marlowe (Greg, 1945; Craig and Kinney, 2009) 
The former is not profiled because his uncontested canon (comprising just \emph{The Spanish Tragedy}) is too small to build a profile and the latter is not well ranked in Acts 1-4 but is close to the second preferred candidate in Act 5. For this reason, we attribute the scenes to either Shakespeare and Marlowe rather than Shakespeare and Jonson. The scene-by-scene analysis shows Shakespeare as the more likely candidate for almost the entire play, with many scenes attributed to Shakespeare by a margin of at least $4cn$. The exceptions to this are Scene 5.5, which is assigned to Marlowe, and Scene 5.2, a tie between candidates. Our results are consistent with existing claims by MacDonald P. Jackson \citep{jackson2006shakespeare} that Shakespeare at the very least wrote the middle of the play (Act 3), and of the candidates tested here he is the most likely to have written Acts 1, 2, and 4 as well. The writer(s) of the non-Shakespearian parts of \emph{Arden of Faversham} may, of course, be person(s) entirely unknown to scholarship, and may include Kyd whom we are unable to test for.

\begin{figure}[h]
\centering
\includegraphics[height=.25\textheight, width=.5\textwidth, keepaspectratio]{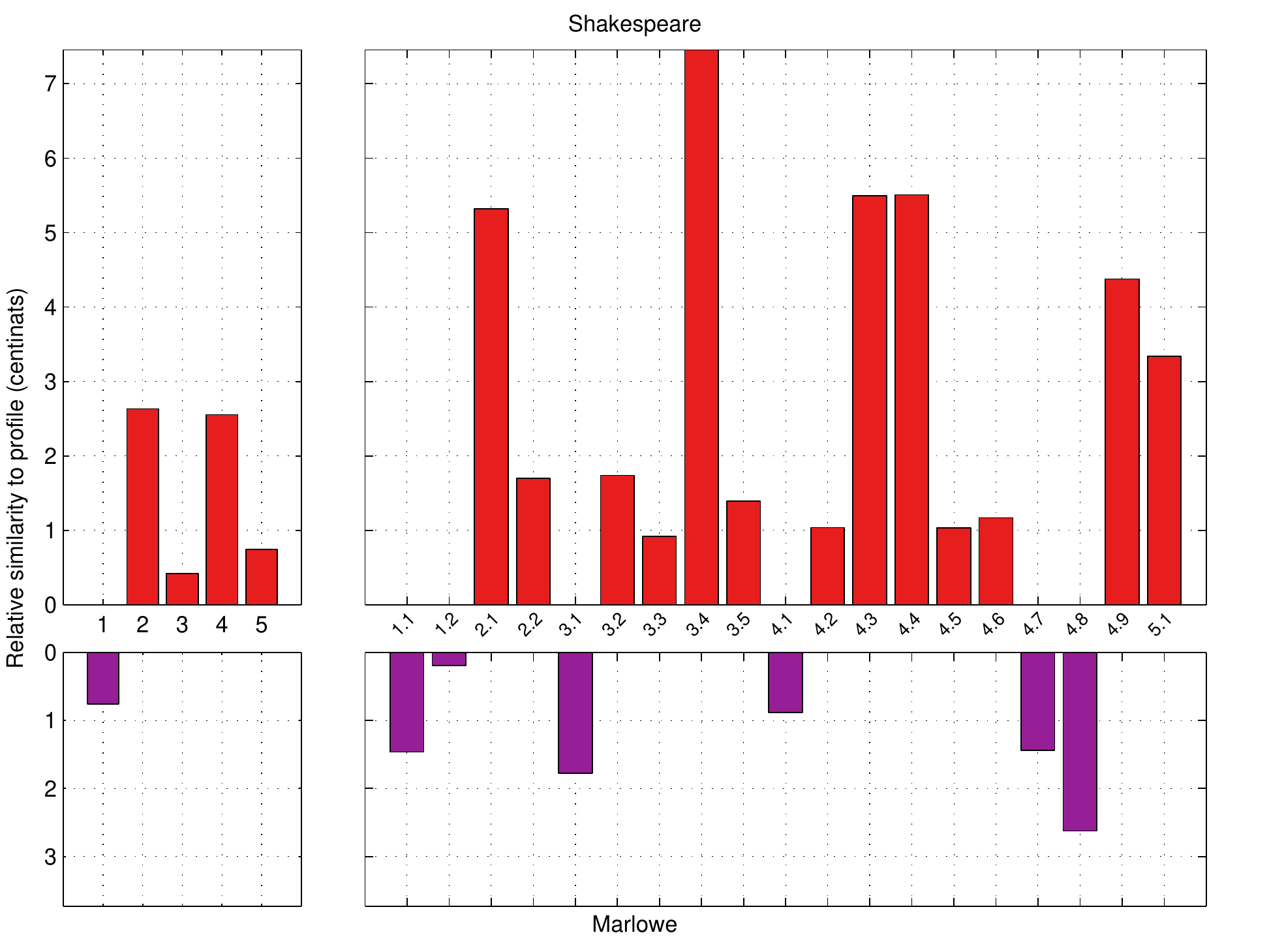}
\caption{Attribution of acts and scenes of \emph{Edward III} between Shakespeare (red) and Marlowe (purple).}
\label{fig_edward_iii_attribution}
\end{figure}

An analysis is additionally performed for \emph{Edward III}. As before, the two most commonly cited candidates for co-authorship with Shakespeare are Kyd and Marlowe (Merriam, 1993; Craig and Kinney, 2009). 
The eight-author act attribution of \emph{Edward III} in Fig. \ref{fig_edward_iii_attribution} shows Act 1 assigned to Marlowe. Acts 2, 4, and 5 are attributed to Shakespeare, as well as Act 3 by a small margin of less than $0.5cn$. A look into the scene-by-scene attribution, however, shows that in addition to 1.1, Marlowe is also assigned Scenes 3.1, 4.1, 4.7, and 4.8, while the analysis of Scene 1.2 does not provide a clear result. While not shown in Fig. \ref{fig_edward_iii_attribution}, the relative entropy values in the attribution of Scene 4.3 is large for both profiles---being at a distance of $+1.5cn$ from Shakespeare and $+7cn$ from Marlowe--- suggesting that neither Shakespeare nor Marlowe, but possibly a third author contributed the scene.

Timothy Irish Watt has suggested that Shakespeare wrote Scenes 1.2 and 2.1 while someone other than Shakespeare, Marlowe, or Peele wrote Scenes 3.1-4.3 \citep{craig2009shakespeare}. Our results point to Shakespeare as a likely candidate for Scene 2.1, with his profile being more than $5cn$ closer than Marlow's profile to the WAN of \emph{Edward III}. Additionally, along with Scene 4.3, we find Scenes 3.2, 3.3, 4.1, 4.2, 4.5, and 4.9 to be possibly written by a third author due to comparatively large distances between the scenes' WANs and the profiles of Shakespeare and Marlowe. Indeed, although not displayed in Fig. \ref{fig_edward_iii_attribution}, the nearest that each of these scenes comes to Shakespeare's or Marlowe's style is between $+0.1cn$ and $+1.7cn$, whereas for all other scenes this distance ranges from $-0.3cn$ to $-3.5cn$.

\subsection{Shakespeare and Peele}

\begin{figure}[h]
\centering
\includegraphics[height=.25\textheight, width=.5\textwidth, keepaspectratio]{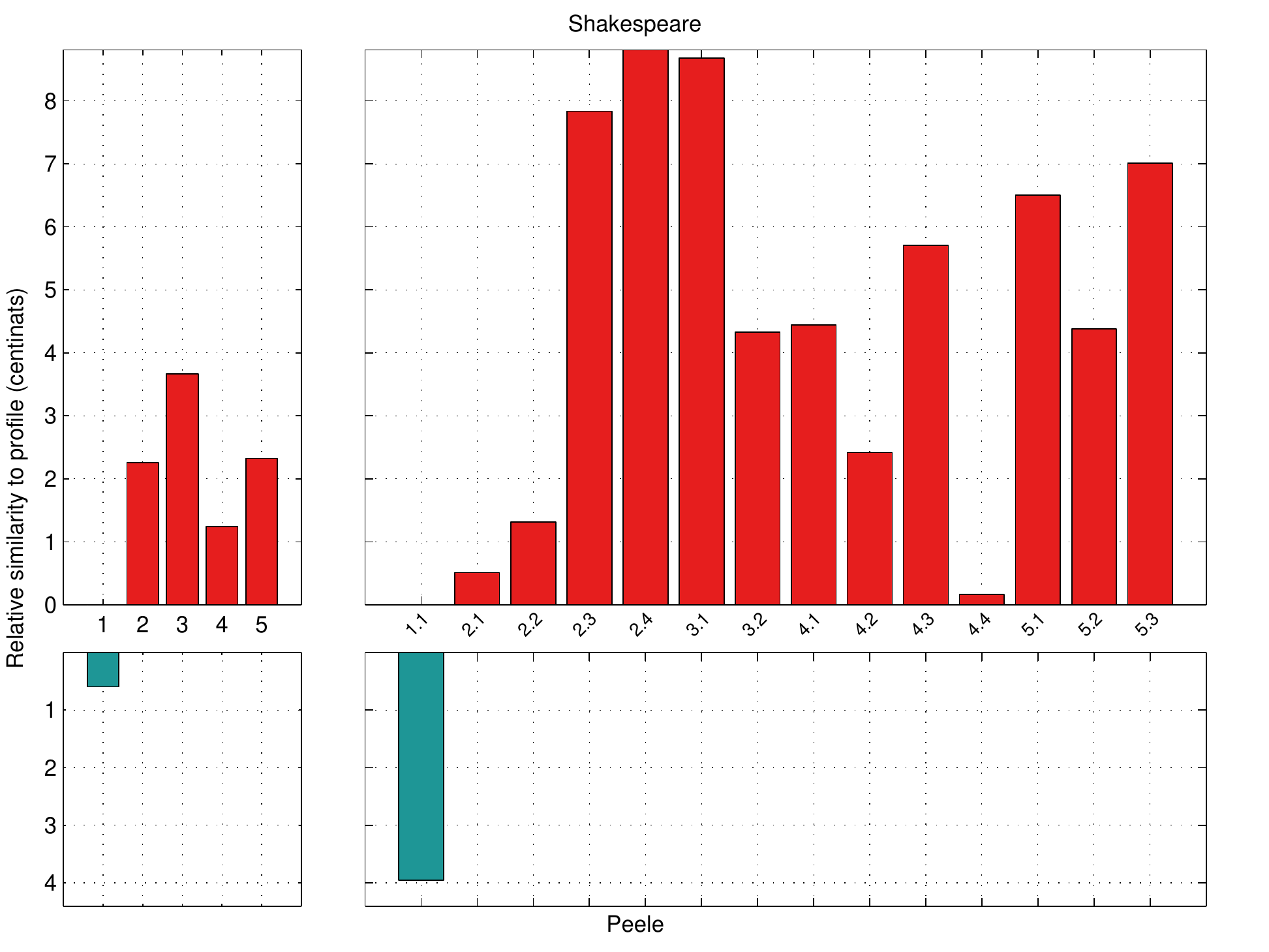}
\caption{Attribution of acts and scenes of \emph{Titus Andronicus}  between Shakespeare (red) and Peele (blue). Note that here the comparative relative entropies for Act 1 and its sole Scene, 1.1, differ. The plot of Scene 1.1 (right) reports the difference in relative entropy between Peele and Shakespeare while the plot of Act 1 (left) reports the difference in relative entropy between Peele and Marlowe, the second ranked author.}
\label{fig_titus_andronicus_attribution}
\end{figure}

%
\begin{table}
\caption{Relative entropies between Scene 3.2 of \emph{Titus Andronicus} and author profiles.}
\centering{\footnotesize
\begin{tabular}{ l  l  l  l  }            \hline
Shakespeare & Fletcher & Jonson & Marlowe  \\
\hline
\textbf{0.47} & 5.69 & 2.76 & \textbf{0.27} \\
\hline

Middleton & Chapman & Peele & Greene \\
\hline
3.72 & 2.73 & 4.80 & 1.12 \\
\hline
\end{tabular}}
\label{table_fly_scene}
\end{table}

Shakespeare's play, \emph{Titus Andronicus}, is generally agreed to be co-authored with Peele \citep{vickers2002shakespeare}, and is attributed act-by-act and scene-by-scene in Fig. \ref{fig_titus_andronicus_attribution}. Act 1 is assigned to Peele while the rest of the play is attributed to Shakespeare. In the scene attributions, Scenes 2.1 and 4.4 are attributed to Shakespeare by a small margin of less than $1cn$. Typical attributions of this play, such as the one performed by Brian Vickers \citep{vickers2002shakespeare}, assign Act 1 to Peele as well as Scenes 2.1 and 4.1. Recently, William W. Weber has cast doubt on Peele's authorship of Scene 4.1 \cite{weber2014shakespeare}, finding strong reasons to give it to Shakespeare, and our method agrees with Weber's conclusion.

The so-called "Fly" scene, 3.2, is present in the 1623 Folio but not in quarto editions, suggesting that it was a later addition to the play and possibly added by another author. The relative entropies for this scene are compared in Table \ref{table_fly_scene}. The two top candidates here are Shakespeare and Marlowe. However, the scene only appeared in editions published long after Marlowe's death so our top candidate for this scene remains Shakespeare. Recently, Middleton has also been proposed as a candidate \citep{taylor2017fly}, however the results in Table \ref{table_fly_scene} do not support this claim.

\section{Conclusion}\label{sec_conclusion}

Function word adjacency networks were used to analyze the authorship of texts written by popular playwrights during the Early Modern English period. WANs were built for a large set of texts in the corpus of the analyzed authors and were compared via a measure of relative entropy. The networks of every text known to be written by a particular author were aggregated to form a profile network. The profile networks were then compared to one another to determine the general similarity between author styles. Each text in an author's corpus was compared to every profile and attributed to the author whose profile network produced the smallest relative entropy. An attribution accuracy of 92.6\% was achieved when attributing amongst all authors. The classification power was then further evaluated with respect to plays written by multiple authors, both through the attribution of an entire play as well as its individual act and scene components. The acts and scenes were individually analyzed in a set of plays with highly disputed co-authorship, in which we both corroborate existing breakdowns and provide evidence of new assignments. We overall find WANs to be simple yet effective tools in distinguishing between playwrights from the Early Modern era by considering relational structures between function words not previously considered in authorship attribution studies from this time period.

\section{Funding}
This work is supported by National Science Foundation CAREER CCF-0952867, National Science Foundation CCF-1217963 and AHRC AH/N007654/1.


\urlstyle{same}
\bibliography{auth_attr_ref}

\newpage

\section*{Appendix}\label{sec_appendix}

\begin{table*}[h]
\caption{Plays used to create sole-authorship canons}
\label{table_profile_texts}
\renewcommand{\arraystretch}{1.2}
\begin{tabular}{c c} 
\input{table_all_texts1.tex} &
\input{table_all_texts2.tex}
\end{tabular}
\end{table*}

\begin{table*}[h]
\centering
\caption{Plays used in co-authorship attributions}
\label{table_att_texts}
\renewcommand{\arraystretch}{1.2}
\begin{tabular}{c c} 
\input{table_att_texts.tex} & 
{\fontsize{7}{7}\selectfont
\begin{tabular}{p{38mm} p{38mm}} 
\hline
\textbf{Shakespeare \& Fletcher}\\\hline
Henry VIII (H8) & Two Noble Kinsmen (TNK)\\
\hline
\textbf{Shakespeare \& Marlowe}        \\\hline
1 Henry VI (1H6) & 
2 Henry VI (2H6) \\
3 Henry VI (3H6) &
Arden of Faversham (ARD) \\
Edward III (E3) & \\
\hline
\end{tabular}}
\end{tabular}
\end{table*}


\begin{table*}[h]
\centering
\caption{Function words used to build WANs. Only the words designated with an $a$ or $s$ are used in the networks used to attribute acts and scenes, respectively.}
\input{table_function_words_all2.tex}
\label{table_function_words_all}
\end{table*}
%


\end{document}

%% file: my_sections.tex
\usepackage{needspace}



%% file: table_profile_distances.tex
\begin{tabular}{l l l l l l l }
\hline
       & Shakespeare     & Fletcher       & Jonson     & Marlowe       & Middleton     & Chapman                \\ \hline
Shakespeare    &    &8.9     &4.7    &8.9    &6.8    &4.8\\
Fletcher    &7.4    &       &7.3    &14.7    &8.0    &8.4\\
Jonson    &4.1    &7.9       &    &11.1    &6.7    &5.4\\
Marlowe    &10.1    &17.4       &13.0    &    &16.5    &12.9\\
Middleton    &5.8    &8.2       &6.3    &14.1    &    &6.6\\
Chapman    &4.7    &9.6       &5.8    &11.4    &7.3    &\\\hline
\end{tabular}

%% file: table_method_comparison.tex

\begin{tabular}{l| l l l l l l }
\hline
\textbf{Method} & WAN & PCA (4 pc's) & PCA (16 pc's) & Delta (Manhattan) & Delta (Euclidean) & Delta (Cosine) \\ \hline
\textbf{Accuracy} & 92.6 & 72.8 & 81.5 & 91.3 & 79.3 & 81.5 \\ \hline
\end{tabular}

%% file: table_method_comparison_scene.tex

\begin{tabular}{l| l l l l l l }
\hline
\textbf{Method} & WAN & PCA (4 pc's) & PCA (16 pc's) & Delta (Manhattan) & Delta (Euclidean) & Delta (Cosine) \\ \hline
\textbf{Accuracy (Act)} & 93.4 &  62.6 & 71.4 & 74.3 & 67.0 & 68.1 \\ 
\textbf{Accuracy (Scene)} & 91.5 & 69.1 & 71.5 & 70.1 & 69.3 & 69.8 \\
\end{tabular}

%% file: table_all_texts1.tex
\renewcommand{\arraystretch}{1.2}
{\fontsize{7}{7}\selectfont
\begin{tabular}{p{38mm} p{38mm}} 
\hline
\textbf{William Shakespeare }       \\\hline
Antony and Cleopatra (ANT) & 
All's Well that Ends Well (AWW) \\
As You Like It (AYL)  &
The Comedy of Errors (ERR) \\
Coriolanus (COR)  & 
Cymbeline (CYM) \\
Hamlet (HAM) & 
1 Henry IV (1H4)  \\
2 Henry IV (2H4) &
Henry V (H5)                   \\
Julius Caesar (JC)              &
King John (JN)	 \\
King Lear (LR)                 &
Love Labour's Lost (LLL)        \\
The Merchant of Venice (MV)     &
The Merry Wives of Winsdor (WIV) \\
A Midsummer Night's Dream (MDB)  &
Much Ado About Nothing (ADO)    \\
Othello (OTH)                   &
Richard II (R2)                 \\
Richard III (R3)                &
Romeo and Juliet (ROM)          \\
The Taming of the Shrew (SHR)    &
The Tempest (TMP)              \\
Troilus and Cressida (TRO)       &
Twelfth Night (TN)              \\
The Two Gentlemen of Verona (TGV) &
The Winter's Tale (WT)          \\ 
\hline

\textbf{Christopher Marlowe}\\\hline
Dr Faustus (DRF) &
Edward II (E2) \\
The Jew of Malta (JEW) &
The Massacre at Paris (MAS) \\
1 Tamburlaine (T1) &
2 Tamburlaine (T2) \\
\hline

\textbf{Ben Jonson} \\\hline
Alchemist (ALC) &
Bartholomew Fair (BAR) \\
Catiline's Conspiracy  (CAT) &
Cynthia's Revels  (CYN) \\
The Devil is an Ass (DIA) &
Epicoene (EPI) \\
Every Man in His Humour (MIH) &
Every Man Out of His Humour (MOH) \\
The Magnetic Lady  (MAG) &
The New Inn (NEW) \\
Poetaster  (POE) &
The Sad Shepherd (SAD) \\
Sejanus's Fall (SEJ) &
The Staple of News (SON) \\
A Tale of a Tub  (TUB) &
Volpone (VOL) \\
\hline
\textbf{George Chapman} \\\hline
All Fools (ALL) &
Sir Giles Goosecap (SGG) \\
Bussy Dambois (BDA) &
Caesar and Pompey (CAP) \\
The Conspiracy of Charles Duke of Byron (CDB) &
The Tragedy of Charles Duke of Byron (TDB) \\
The Gentlemen Usher (GEN) &
A Humorous Day's Mirth (HDM) \\
May Day (MAY)&
Monsieur D'Olive (MDO) \\
The Blind Beggar of Alexandria (BBA) & 
The Revenge of Bussy Dambois (RBD)  \\
The Widow's Tears (WID)  \\
\hline
\end{tabular}}

%% file: table_all_texts2.tex
\renewcommand{\arraystretch}{1.2}
{\fontsize{7}{7}\selectfont
\begin{tabular}{p{38mm} p{38mm}} 
\hline
\textbf{John Fletcher} \\\hline
Bonduca (BON) &
Chances (CHA) \\
The Faithful Shepherdess (TFS) &
The Humorous Lieutenant (HUM) \\
The Island Princess (ISL)  &
The Loyal Subject (LOY) \\
The Mad Lover (TML)  &
Monsieur Thomas  (THO)\\
The Pilgrim (PIL)  &
Rule a Wife and Have a Wife (RAW) \\
Valentinian (VAL)  &
Wife for a Month (WFM) \\
The Wild Goose Chase  (WGC)  &
The Woman's Prize (WPR) \\
Women Pleased (WPL) \\
\hline
\textbf{Fletcher \& Francis Beaumont}     \\
   \hline
The Coxcomb (COX) &
Cupid's Revenge (CUP)\\
A King and No King (KNK) &
The Maid's Tragedy (TMT)\\
Philaster (PHI) &
The Scornful Lady (TSL)\\
The Woman Hater (TWH) &
Love's Pilgrimage (PIL)\\

\hline
\textbf{Fletcher \& Phillip Massinger}   \\    
 \hline
The Custom of the Country (COC) &
The Double Marriage  (TDM) \\
The Elder Brother (TEB) &
The False One (TFO) \\ 
John Van Olden Barnavelt (JVO) & 
The Little French Lawyer (LFL)  \\
The Prophetess  (PRO) &
The Sea Voyage (SEA) \\
Spanish Curate (TSC) & 
A Very Woman (TVW)  \\
\hline
\textbf{Thomas Middleton} \\\hline
Your Five Gallants (FIV) &
A Game at Chess  (GAC)\\
A Mad World My Masters (MAD) &
A Chaste Maid in Cheapside  (MAC) \\
Hengist King of Kent (HEN) &
Michaelmas Term (MIC) \\
More Dissemblers Besides Women (DIS) &
No Wit, No Help Like a Woman's (NOW)\\
The Phoenix  (PHO) &
The Puritan Widow (PUR) \\
The Revenger's Tragedy  (REV) &
The Second Maiden's Tragedy (SMT) \\
A Trick to Catch the Old One (TCO) &
The Widow (WID) \\
The Witch (WTH) &
Women Beware Women (BEW) \\
\hline
\textbf{Robert Greene}       \\
\hline
Friar Bacon and Friar Bungay & 
Orlando Furioso  \\ 
James IV &
Alphonsus, King of Aragon \\
\hline
\textbf{George Peele}        \\
\hline
The Arraignment of Paris &
Edward I \\ 
The Battle of Alcazar &
The Love of King David and Fair Bathsheba \\ 
Old Wives Tale \\
\hline
\end{tabular}}

%% file: table_att_texts.tex
{\fontsize{7}{7}\selectfont
\begin{tabular}{p{38mm} p{38mm}} 
\hline

\textbf{Jonson \& Chapman} \\\hline
Eastward Ho (HO) & \\
\hline

\textbf{Shakespeare \& Middleton} \\\hline
Macbeth (MAC) &
Measure for Measure (MEA) \\
Timon of Athens (TIM) & \\
\hline


\end{tabular}}

%% file: table_function_words_all2.tex
\begin{tabular}{llllllllll}
$\text{a}^{as}$ 		& $\text{at}^{as}$		& $\text{could}^{as}$		& $\text{in}^{as}$	& $\text{much}^{a}$		& $\text{off}^{as}$	& $\text{past}^{as}$	& $\text{them}$			& $\text{to}^{as}$		& $\text{while}^{a}$	\\
$\text{about}^{a}$	& $\text{away}^{a}$		& $\text{dare}^{as}$		& $\text{into}^{as}$	& $\text{must}^{a}$		& $\text{on}^{as}$	& $\text{shall}$		& $\text{then}$			& $\text{until}^{as}$		& $\text{who}^{as}$	\\
$\text{after}$		& $\text{bar}^{as}$		& $\text{down}^{as}$		& $\text{it}^{as}$	& $\text{need}^{a}$		& $\text{once}^{as}$	& $\text{should}^{a}$	& $\text{therefore}^{as}$	& $\text{unto}$			& $\text{whom}^{as}$	\\
$\text{against}^{as}$	& $\text{because}^{as}$	& $\text{enough}^{a}$	& $\text{like}$		& $\text{neither}$		& $\text{one}$		& $\text{since}^{as}$	& $\text{these}^{a}$		& $\text{up}^{as}$		& $\text{whose}^{as}$	\\
$\text{all}^{as}$		& $\text{before}^{a}$		& $\text{every}^{as}$		& $\text{little}$		& $\text{next}^{a}$		& $\text{or}^{as}$	& $\text{so}$		& $\text{they}^{as}$		& $\text{upon}^{a}$		& $\text{will}^{a}$	\\
$\text{an}^{a}$		& $\text{both}$			& $\text{for}^{a}$		& $\text{many}$	& $\text{no}^{a}$		& $\text{other}^{as}$	& $\text{some}^{as}$	& $\text{this}^{as}$		& $\text{us}$			& $\text{with}^{as}$	\\
$\text{and}^{as}$	& $\text{but}$			& $\text{from}$			& $\text{may}^{as}$	& $\text{none}^{a}$		& $\text{our}^{as}$	& $\text{such}$		& $\text{those}^{as}$		& $\text{what}$			& $\text{within}^{as}$	\\
$\text{another}^{as}$	& $\text{by}^{as}$		& $\text{given}^{as}$		& $\text{might}^{as}$	& $\text{nor}$			& $\text{out}^{as}$	& $\text{than}^{as}$	& $\text{though}^{a}$	& $\text{when}$			& $\text{without}^{as}$	\\
$\text{any}^{as}$	& $\text{can}^{a}$		& $\text{hence}^{as}$	& $\text{more}^{as}$	& $\text{nothing}^{as}$	& $\text{over}^{as}$	& $\text{that}^{as}$	& $\text{through}^{as}$	& $\text{where}^{a}$		& $\text{would}^{as}$	\\
$\text{as}$		& $\text{close}^{a}$		& $\text{if}^{as}$		& $\text{most}$		& $\text{of}$			& $\text{part}$		& $\text{the}^{as}$	& $\text{till}$			& $\text{which}^{as}$	& $\text{yet}^{as}$
\end{tabular}